\documentclass[journal]{IEEEtran}
\usepackage{amsmath,amsfonts}
\usepackage{amssymb}
\usepackage{algorithmic}
\usepackage{algorithm}
\usepackage{array}
\usepackage[caption=false,font=normalsize,labelfont=sf,textfont=sf]{subfig}
\usepackage{textcomp}
\usepackage{stfloats}
\usepackage{url}
\usepackage{verbatim}
\usepackage{graphicx}
\usepackage{adjustbox}
\usepackage{cite}
\usepackage{epstopdf}
\usepackage{listings}
\usepackage{xcolor}
\usepackage{hyperref}

\hypersetup{colorlinks=true, allcolors=blue}

\usepackage{cancel}

\begin{document}

\title{Learning Legged MPC with Smooth Neural Surrogates} 
\author{%
Samuel A. Moore$^1$, Easop Lee$^2$, and Boyuan Chen$^{1,2,3}$%
\thanks{$^1$Department of Mechanical Engineering and Materials Science, Duke University, Durham, NC, USA.}%
\thanks{$^2$Department of Electrical and Computer Engineering, Duke University, Durham, NC, USA.}%
\thanks{$^3$Department of Computer Science, Duke University, Durham, NC, USA.}%
}





\maketitle

\IEEEpeerreviewmaketitle

\begin{abstract}
Learning and model predictive control can be complementary for legged robotics, but integrating learned dynamics with online planning remains difficult. Neural network dynamics suffer from three issues: stiff transitions inherited from the training data, non-physical local nonsmoothness, and non-Gaussian model errors from rapid state changes. We address the first two with the smooth neural surrogate, a neural network with tunable smoothness that provides informative predictions and well-conditioned derivatives for trajectory optimization through contact. To address the third, we train these models using a heavy-tailed likelihood that we show to be a better match for empirical error distributions. Together, these choices substantially improve learned model predictive control. Across zero-shot locomotion tasks of increasing difficulty, smooth neural surrogates reduce cumulative cost on well-conditioned behaviors (typically 10–50\%) and yield substantially larger gains in regimes where standard neural dynamics fail. In these regimes, smoothing enables reliable execution (from 0/5 to 5/5 success) and $\approx$2–50$\times$ lower cumulative cost, reflecting vast improvements in controller generalization.
\end{abstract}

\section{Introduction}

\IEEEPARstart{R}{einforcement} learning (RL) and model predictive control (MPC) have both seen widespread success in legged robotics. Learning provides strong representational flexibility and scalability, particularly through techniques such as domain randomization, enabling capabilities like blind locomotion over challenging terrain~\cite{lee2020learning, kumar2021rma}. Model predictive control, by contrast, enables online behavior shaping by adjusting costs and constraints, making it well suited for rapid task switching without retraining~\cite{tassa2012synthesis, howell2022predictive}. Despite these complementary strengths, the effectiveness of MPC within an RL pipeline remains limited for legged robots, and existing controllers tend to favor either learned policies with limited task-level flexibility or analytic models that are too scenario-specific to enable robust control across many environments.

Learned MPC, which combines learned dynamics with online planning, offers a promising middle ground. However, improving the open-loop prediction error of neural dynamics does not guarantee improved control performance~\cite{lutter2021learning}. For online MPC, the model is not only a simulator: it must be compact enough for repeated real-time rollouts and differentiation, and smooth enough to provide well-conditioned local derivatives. Even exact analytic models can perform poorly when contact dynamics are not handled appropriately within the optimization~\cite{posa2013direct, kelly2017introduction, kim2025contact}.

\begin{figure}[!t]
    \centering
    \includegraphics[width=\linewidth]{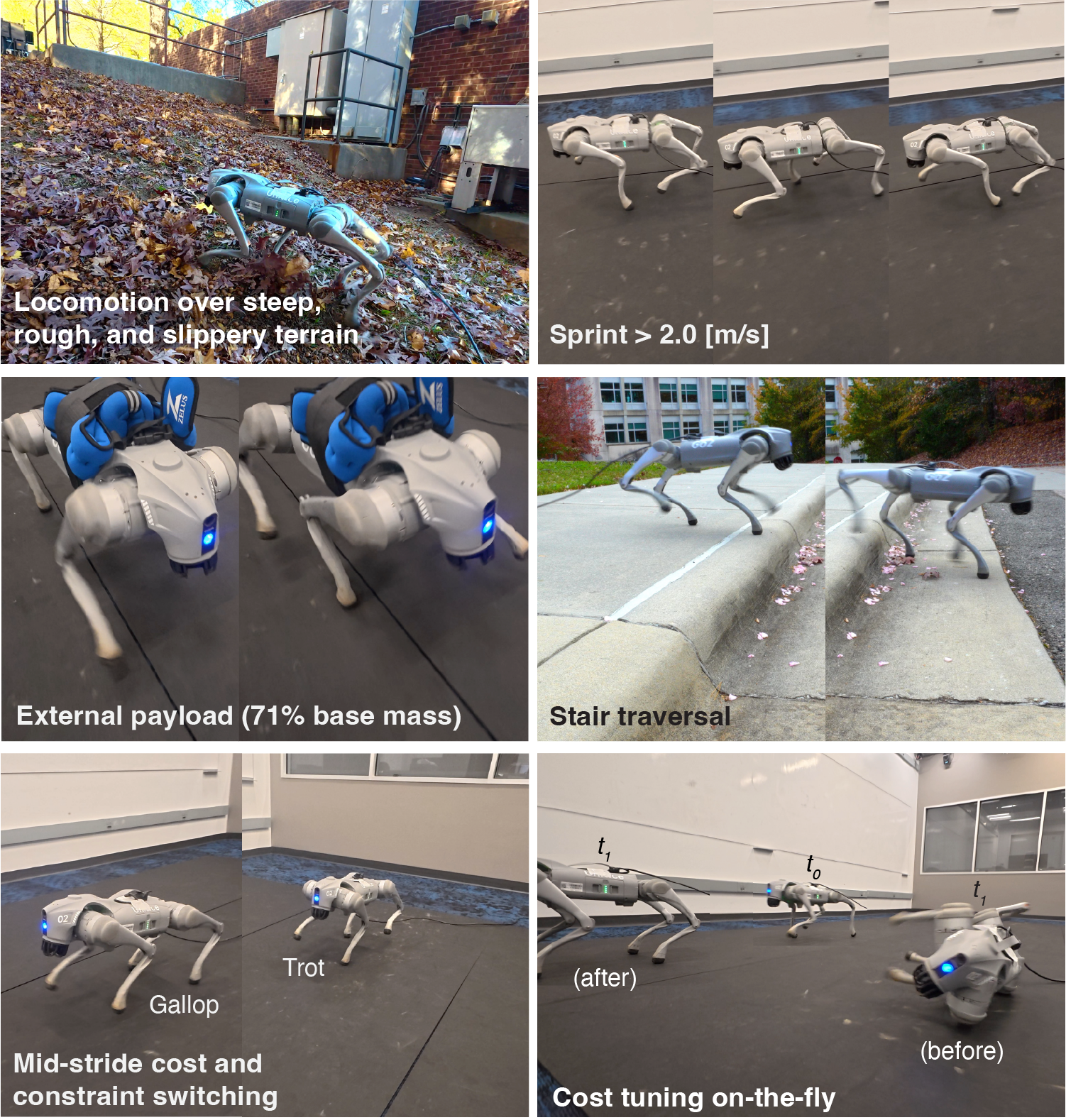}
    \caption{\textbf{Versatile quadruped control with smooth neural dynamics. a)} Our smooth neural surrogates trained with heavy-tailed losses learn terrain-varying dynamics and state estimation while generalizing to new tasks outside the training domain. Together, they combine the representational flexibility of deep learning with the task-level adaptability of MPC, enabling reliable whole-body control across diverse behaviors and environments.}
    \label{fig: teaser}
\end{figure}

\begin{figure}
    \centering
    \includegraphics[width=\linewidth]{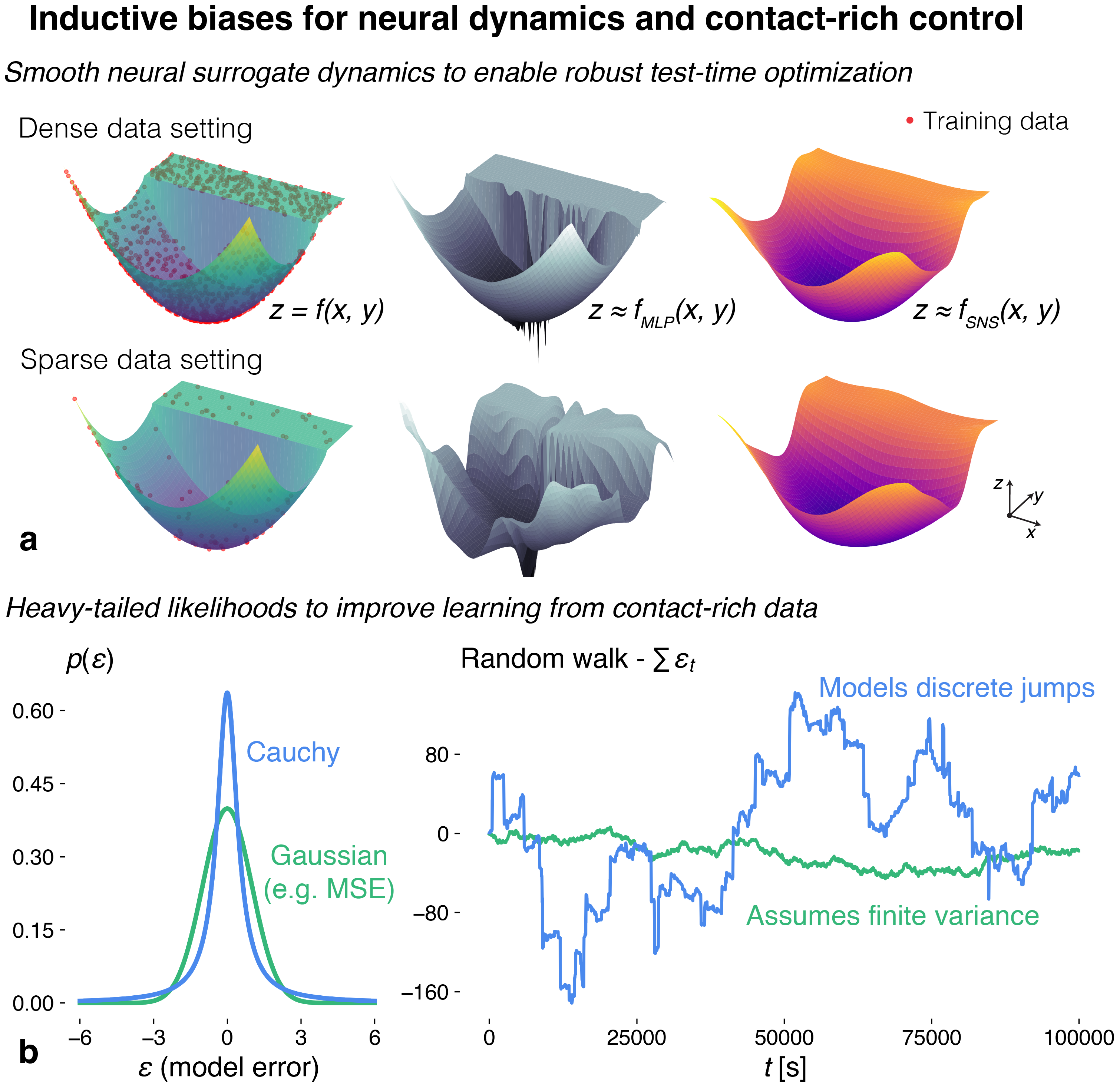}
    \caption{\textbf{Improving model learning for contact-rich control. a}, Smooth neural surrogates, structured with Lipschitz-based weight normalization, improve the generalization of learned MPC. These networks reduce both physical nonsmoothness from contact events and local nonsmoothness related to overfitting. As underlying dynamics may be nonsmooth, these smooth neural surrogates bias the learned representation to improve trajectory optimization as opposed to physical accuracy. \textbf{b}, Commonly used Gaussian likelihoods for dynamics learning, like the MSE, fail to capture discrete jumps in the model error. In contrast, Cauchy likelihoods naturally describe impulsive contacts, improving model learning from stiff, contact-rich data.}
    \label{fig: high-level idea}
\end{figure}

\begin{figure*}
    \centering
    \includegraphics[width=\linewidth]{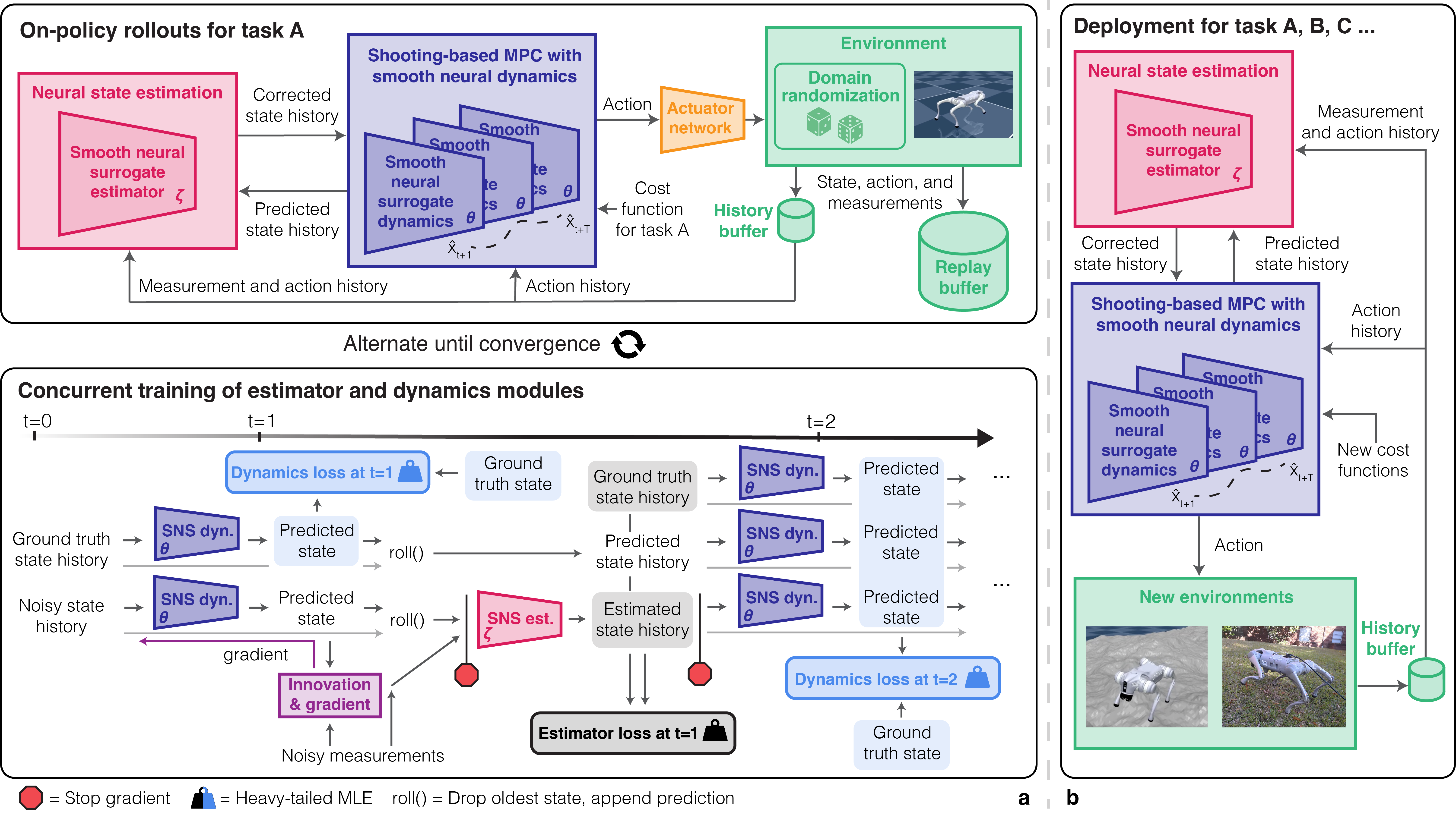}
    \caption{\textbf{Our framework for learning generalizable legged MPC and state estimation. a}, We alternate data collection and model training update steps until convergence. We define the task as a combination of reward or cost function and environment. Data is collected while attempting to track a desired body velocity or orientation and stored in a replay buffer. The smooth neural surrogate state estimator and dynamics modules are trained concurrently. \textbf{b}, As a result of the training procedure, the framework can generalize to new tasks at deployment.}
    \label{fig: method fig}
\end{figure*}
Our starting point is a simple but important observation: \emph{the pathologies of optimization under contact dynamics, particularly for gradient-based methods, are compounded when the dynamics are modeled with standard neural networks.} We hypothesize that these failures are driven primarily by nonsmoothness, which manifests in three ways:
\begin{enumerate}
\item Neural network dynamics often inherit the very stiff transitions present in the ground-truth contact dynamics, making contacts difficult to optimize through.
\item Neural networks frequently introduce additional local nonsmoothness and, consequently, local optima that do not exist in the underlying physical dynamics.
\item Training data for these stiff dynamics necessarily contain rapid state changes, which can destabilize learning and violate common Gaussian model-error assumptions.
\end{enumerate}

A promising direction for addressing 1) and 2) is to explicitly control neural network smoothness. Smooth architectures and Lipschitz-constrained networks can yield better gradients for test time optimization, improved generalization, and robustness to perturbations \cite{liu2022learning, rosca2020case, anil2019sorting, gulrajani2017improved, yoshida2017spectral}. Yet, these ideas have seen limited use in control and robotics \cite{song2023lipsnet, tan2024robust}, especially for learning stiff legged robot dynamics. 


 Addressing 3) requires checking the assumptions behind the loss function and the implied likelihood model. In practice, Gaussian error assumptions are common \cite{xu2025neural, lutter2021learning, chua2018deep, amigo2025first, byravan2021evaluating, parmar2021fundamental, roth2025learned}, but they are often unverified and inappropriate for datasets that introduce heavy-tailed errors, a phenomenon we explicitly observe in legged-robot dynamics. Cauchy noise models are known to perform significantly better when the errors are impulsive in nature~\cite{tsakalides1994maximum}.

We study these pathologies using full-order single-shooting MPC with learned
dynamics and state estimation. Single shooting is the simplest form of trajectory optimization: it parameterizes an open-loop
control sequence, rolls the dynamics forward, and updates the controls to minimize a cost. Although it can be paired with general-purpose optimizers \cite{nocedal2006numerical}, gradient-based single shooting is sensitive to poor conditioning, local minima, and nonsmoothness~\cite{kelly2017introduction, kelly2017transcription}. Sampling-based MPC is more tolerant of nonsmoothness \cite{xue2025full, howell2022predictive, chua2018deep, lutter2021learning, lee2025sym2real, nagabandi2020deep}, but scales poorly with action dimension and still depends on the quality of the learned dynamics.

These observations motivate our central question: can smooth neural dynamics and heavy-tailed likelihoods produce compact learned models that are accurate and regular enough for gradient-based MPC through contact?

We answer this question by extending smooth neural architectures for use as learned dynamics models in MPC through contact. We introduce the \emph{smooth neural surrogate} (SNS): a multilayer perceptron with learned layer-wise Lipschitz constants and tunable smoothness (Fig.~\ref{fig: high-level idea}a, Section~\ref{sec: smooth neural networks}). We pair this architecture with robust maximum-likelihood estimation using heavy-tailed error models, which better match the empirical error distribution observed in legged-robot dynamics (Fig.~\ref{fig: high-level idea}b, Section~\ref{sec: dynamics and estimation}). We focus on multilayer perceptrons (MLPs) as a controlled setting to isolate the effects of smoothness and robust learning on trajectory optimization, rather than on maximizing long-horizon open-loop simulation fidelity.


We train both the dynamics and a state estimator under domain randomization, and deploy the resulting models in an MPC framework (Section~\ref{sec: results}). These learned modules are then transferred to hardware and evaluated on unseen terrains and tasks without additional training (Fig.~\ref{fig: teaser}, Section~\ref{sec: results}).

\section{Overview and Contributions}
\noindent Fig.~\ref{fig: method fig} provides a high-level overview of our approach to learned, gradient-based MPC for legged robots under domain randomization and partial observability. An SNS-MLP state estimator refines prior state estimates using corrections informed by the learned dynamics. The SNS-MLP dynamics model, in turn, predicts future states based on the refined estimates. Both modules are trained using heavy-tailed (Cauchy) maximum-likelihood estimation. The estimator’s smoothness acts primarily as a regularizer, reducing perturbations to the downstream MPC problem.  The dynamics model’s smoothness is explicitly designed to stabilize gradient-based estimation and single-shooting MPC through contact.

Our primary technical contributions are:
\begin{enumerate}
    \item  We introduce the smooth neural surrogate (SNS-MLP), a neural network with a tunable Lipschitz bound and an exponential parameterization of layerwise smoothness constants that stabilizes learning (Section~\ref{sec: smooth neural networks}). We show that this architecture yields more informative predictions and derivatives for legged MPC through contact than standard and Lipschitz MLP baselines (Section~\ref{sec: results}).

    \item We demonstrate that dynamics residuals in model-based reinforcement learning for legged robots can exhibit heavy-tailed behavior (Section~\ref{sec: dynamics and estimation}), and show that heavy-tailed likelihoods significantly improve training convergence and downstream control performance (Section~\ref{sec: results}).

    \item We show that, when paired with appropriately smooth learned dynamics, gradient-based single-shooting MPC can outperform strong sampling-based methods in scalability, convergence, and efficiency, while the same smoothing also improves the performance of sampling-based MPC (Section~\ref{sec: results}).

    \item We learn full-order dynamics and state estimation under domain randomization (Section~\ref{sec: dynamics and estimation}) and deploy the resulting models on a real quadruped robot, synthesizing new behaviors in unseen environments without additional training (Section~\ref{sec: results}).
\end{enumerate}

\section{Related Work}

\subsection{Model-Based Control of Hybrid Systems}
\noindent Contact dynamics are hybrid, with smooth continuous modes separated by discrete, often non-differentiable transitions. Existing model-based controllers typically handle this difficulty either by avoiding derivatives, exploiting contact structure, or smoothing the dynamics to recover useful gradients.

Sampling-based single-shooting methods avoid derivative computation and are attractive because of their simplicity and GPU parallelism. Recent work has deployed such methods on quadrupeds with full-order rigid-body models~\cite{alvarez2025real, xue2025full}, often reducing the search dimension with spline-based action parameterizations~\cite{howell2022predictive, alvarez2025real, xue2025full}. However, sampling-based MPC remains computationally expensive, scales poorly with action dimension, and is still sensitive to model regularity. Well-conditioned gradient-based methods can offer faster convergence and better scalability when informative derivatives are available~\cite{nocedal2006numerical, sun2018evolving}.

Gradient-based optimization for legged robot control is a broad area; see Wensing \emph{et al.}~\cite{wensing2024optimization} for a recent survey. Contact-implicit trajectory optimization attempts to ensure that contact interactions can be resolved without prescribing a mode sequence in advance~\cite{posa2013direct, le2024fast, kim2025contact, zhang2025whole, mordatch2012discovery, erez2012trajectory}. Some reduce online computation through tailored solver accuracy, cascaded model fidelity, or specialized shooting and whole-body-control pipelines~\cite{khazoom2024tailoring, li2025cafempc}. Another line avoids differentiating through contact by separating contact reasoning from continuous motion optimization: contact phases, timings, locations, or schedules are made explicit, and the remaining centroidal or kino-dynamic problem is solved with structured gradient-based, convexified, or MPC methods~\cite{ponton2021efficient, meduri2023biconmp}. Some contact-implicit methods smooth contact dynamics so that gradients remain useful for planning. Differentiable simulators do this by relaxing complementarity constraints or regularizing contact forces~\cite{le2024fast, kim2025contact, howell2022dojo, zhang2025whole, mordatch2012discovery, erez2012trajectory}. These approaches demonstrate dynamic legged behaviors using analytic models and explicit contact structure; our work is complementary, studying learned full-order dynamics models in which contact effects are embedded in the model and the resulting derivatives must remain useful for MPC.

More general randomized smoothing methods can be applied to arbitrary models, including neural networks, but require additional samples and introduce stochastic gradient noise~\cite{suh2022bundled, pang2023global, howell2022dojo}. Similar smoothing principles have also improved policy learning through differentiable simulation by regularizing contact forces or truncating planning horizons~\cite{xu2022accelerated, georgiev2024adaptive, schwarke2025learning}. Our work follows the smoothing perspective, but learns the smooth surrogate dynamics directly from data rather than imposing smoothing through an analytic simulator or contact model.

\subsection{Neural Control and Zero-Shot Generalization}
\noindent Learned control methods use neural models as dynamics, or policy representations. Most model-based approaches still rely on policy gradients, value functions, or sampling-based planners~\cite{amigo2025first, hafner2019dream, li2025robotic, chua2018deep, lutter2021learning, williams2017information, lee2025sym2real, hu2025egocentric}. Model-free methods are effective, but are often poorly suited for zero-shot task-level adaptation and convergence of sampling-based optimization is typically heuristic and sensitive to hyperparameters. Diffusion-based planners offer another route to flexible behavior generation~\cite{janner2022planning}, but for quadruped control still rely on model-free components or demonstrations~\cite{huang2025flexible}.

Forward-backward representation learning and behavioral foundation models, such as BFM-Zero~\cite{li2025bfmzero}, provide zero-shot adaptation through offline-learned representations and policy classes. This differs from online MPC: deployment-time objectives must be expressible in the learned feature basis and realizable by the trained policy distribution, whereas MPC keeps costs and constraints explicit and optimizes them directly through a dynamics model. Existing demonstrations are built around humanoid control, motion-capture-regularized pretraining, and controlled deployment settings, which do not directly address platforms without large-scale motion capture data, or outdoor deployment. Our focus is therefore compact learned dynamics that can be repeatedly rolled out and differentiated inside real-time MPC.

Other works add contact-aware or physics-informed priors~\cite{pfrommer2021contactnets, parmar2021fundamental, hochlehnert2021learning, lutter2019deep}, reduced-order models and fixed low-level controllers~\cite{roth2025learned, sukhija2022gradient}, or hybrid simulation-learning pipelines~\cite{xu2025neural, heiden2021neuralsim}. High-capacity neural simulators target open-loop simulation fidelity, while our setting prioritizes compact models with low-latency rollouts, inexpensive derivatives, and well-conditioned local gradients for MPC. Smooth neural surrogates and heavy-tailed MLE are compatible with these formulations when learned dynamics are used as optimization models rather than offline data generators.

\subsection{Smooth Neural Networks and Robust Optimization}
\noindent Bounding a network’s Lipschitz constant has been shown to improve adversarial robustness and generalization~\cite{liu2022learning, yoshida2017spectral, miyato2018spectral, gouk2021regularisation}. However, many smoothing techniques are either computationally expensive or overly restrictive~\cite{gulrajani2017improved, rosca2020case}. Strict $1$-Lipschitz methods based on weight orthonormalization~\cite{anil2019sorting} can be under-expressive and unstable in practice, while spectral normalization methods are approximate, hyperparameter-sensitive, and incur nontrivial overhead~\cite{yoshida2017spectral, miyato2018spectral}. Liu \emph{et al.} introduced the Lipschitz MLP to address these issues using weight normalization~\cite{liu2022learning}. Our SNS-MLP extends this approach with a reparameterization that improves convergence and training stability.

In trajectory optimization, Lipschitz-constrained networks have primarily been evaluated in benign, smooth, and low-dimensional settings~\cite{tan2024robust}, leaving their effectiveness for high-dimensional, stiff legged-robot dynamics largely unexplored. Although Lipschitz-bounding techniques exist for more expressive architectures~\cite{qi2023lipsformer}, we focus on standard MLPs as a controlled setting to isolate the effects of smoothness.

Finally, robust optimization literature has long recognized that the assumptions of Gaussian losses, light-tailed errors and finite variance, are frequently violated in practice~\cite{mlotshwa2022cauchy, tsakalides1994maximum}. When residuals are heavy-tailed or impulsive, Gaussian losses overweight outliers, motivating robust alternatives such as Cauchy or Huber losses that reduce their influence~\cite{mlotshwa2022cauchy, tsakalides1994maximum, el2009robust, barron2019general}. Despite this, Gaussian error models remain standard when learning dynamics for systems with contact~\cite{lutter2021learning, nagabandi2020deep, chua2018deep, amigo2025first}.

\section{Smooth Neural Representations}\label{sec: smooth neural networks}
\subsection{Preliminaries}
\noindent \textbf{Lipschitz continuity.} Many general-purpose optimization algorithms rely on \emph{Lipschitz continuity} as a key ingredient in local convergence guarantees and, in practice, its importance to algorithm performance is well known \cite{nocedal2006numerical}. A function $f$ is $c$-Lipschitz if small input changes produce proportionally bounded output changes:
\begin{equation}
\| f(a_0) - f(a_1) \|_p \;\le\; c \, \| a_0 - a_1 \|_p \quad \forall a_0, \,a_1 
\end{equation}
where $\|\cdot\|_p$ is the $p$-norm and $c \ge 0$ is the \emph{Lipschitz constant}. 

\noindent \textbf{Smooth neural networks.} There is no general formula for directly measuring or shaping the Lipschitz constant of an arbitrary analytic function. However, smooth surrogates based on convolution can make a discontinuous function continuously differentiable, providing some qualitative control over smoothness but at the cost of requiring many additional samples at each evaluation point \cite{suh2022bundled, pang2023global}. 

Neural network architectures, by contrast, follow fixed analytic templates while remaining capable of approximating arbitrary functions, making quantitative assessments of smoothness more tractable. Consider an $L$-layer feedforward MLP
\begin{equation}
f(x)=h^{(L)}\circ \cdots \circ h^{(1)}(x),
\quad
h^{(\ell)}(z)=\varphi\big(W^{(\ell)}z+b^{(\ell)}\big),
\end{equation}
For our analysis, we assume that the activation, $\varphi$, is globally Lipschitz and satisfies
\begin{equation}
\|\varphi(a_0)-\varphi(a_1)\|_p \le c_\varphi \|a_0-a_1\|_p
\qquad \forall a_0,a_1 .
\end{equation}
Under this assumption, the Lipschitz constant of each layer satisfies
\begin{equation}
\text{Lip}(h^{\ell})
\le
c_\varphi \,\|W^{(\ell)}\|_p .
\end{equation}
Here, $\|W\|_p$ denotes the induced matrix norm subordinate to the vector $p$-norm, i.e.,
\begin{equation}
\|W\|_p = \sup_{x\neq 0}\frac{\|Wx\|_p}{\|x\|_p}.
\end{equation}
Then, by composition the full network satisfies
\begin{equation}
\label{eq:c_mlp}
c_{\text{MLP}}
\;\lesssim\;
\prod_{\ell=1}^L \|W^{(\ell)}\|_p ,
\end{equation}
with proportionality determined by the activation functions; for $1$-Lipschitz activations (e.g., softplus, tanh), the relation becomes exact. Thus, for neural networks, a direct and efficient way to quantify and shape smoothness is through a Lipschitz upper bound rather than through post hoc smoothing. Importantly, however, \eqref{eq:c_mlp} is generally conservative and may substantially overestimate the \emph{true} Lipschitz constant of the realized network. Accordingly, the budgets used throughout the paper should be interpreted as conservative upper-bound budgets on worst-case sensitivity, rather than tight characterizations of the realized network smoothness. 

Although conservative, the bound in Eq.~\eqref{eq:c_mlp} still provides meaningful quantitative control over global smoothness and removes the need for randomized smoothing. Compared to naive penalization of Eq.~\eqref{eq:c_mlp}, weight normalization strategies have proven more fruitful \cite{liu2022learning}. The bounds in this paper are stated for feedforward MLPs and extensions to residual, convolutional, or attention-based architectures would require architecture-specific norm bounds.

\begin{figure}[t]
    \centering
    \includegraphics[width=0.90\linewidth]{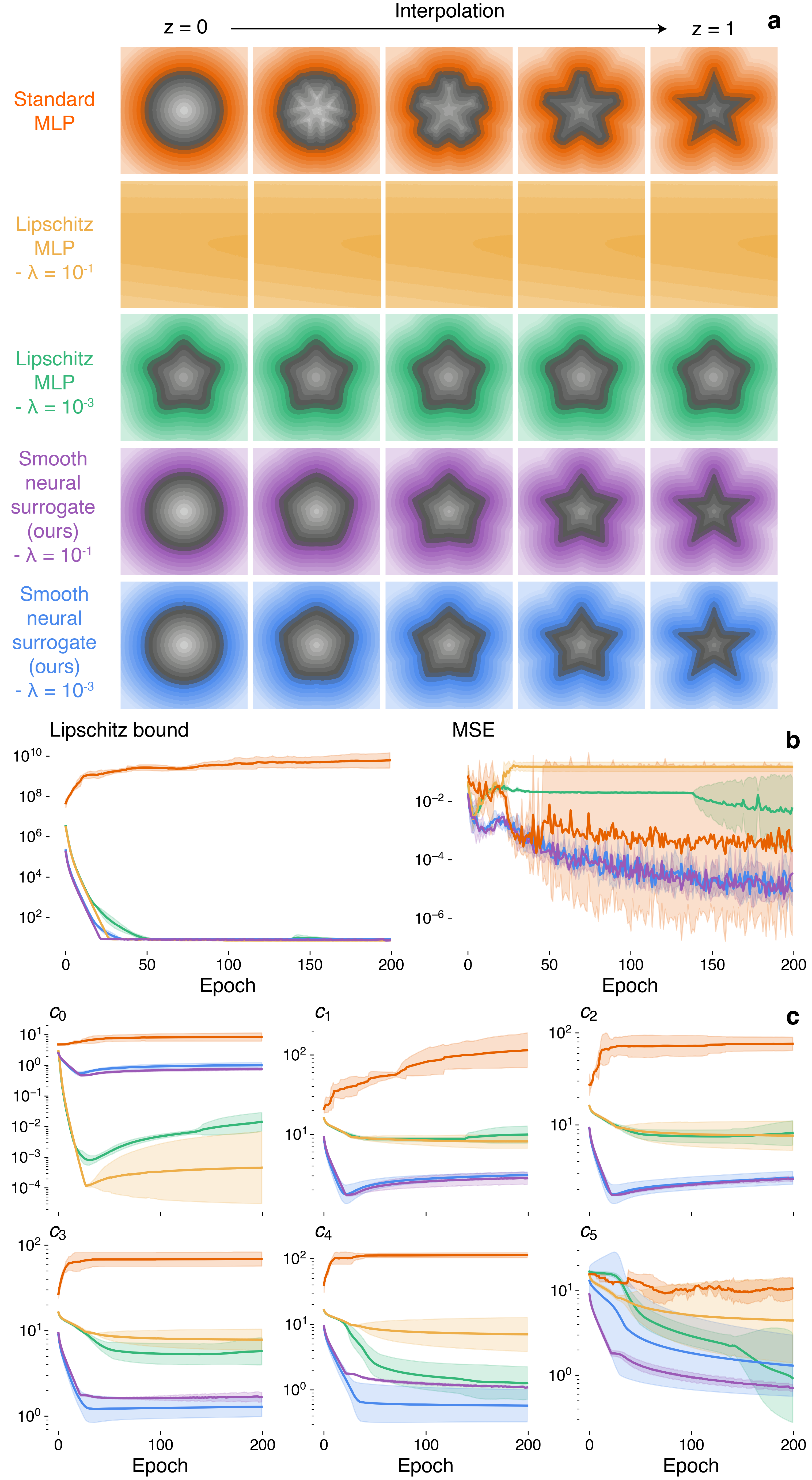}
    \caption{\textbf{Smooth neural surrogates converge under strict Lipschitz 
    constraints and enable zero-shot generalization. a}, Lowest-MSE models for 
    the 2-D shape interpolation task. Only the SNS converges 
    under tight Lipschitz budgets; the Lipschitz MLP \cite{liu2022learning} 
    collapses whether $\lambda=10^{-1}$ or $\lambda=10^{-3}$. At $\lambda=10^{-1}$, the Lipschitz MLP fails to learn the shapes, so its panels show no surface. \textbf{b}, Evolution of Lipschitz upper bounds and training MSE. \textbf{c}, Collapse of the Lipschitz MLP coincides with asymmetric shrinkage of the layerwise Lipschitz constants, consistent with the analysis in Section~\ref{sec: smooth neural networks} and Appendix~\ref{ap:softplus-exp-proof}.
    }
    \label{fig:shape interpolation}
\end{figure}

\subsection{Smooth Neural Surrogates}

\noindent \textbf{Lipschitz-based weight normalization.} We base our approach on Liu \emph{et al.} \cite{liu2022learning}, who use a weight normalization layer paired with learned scalars $c_\ell$ that parameterize the Lipschitz constant of each layer by scaling the $\infty$-norm of $W^{\ell}$:
\begin{equation}
\label{eq:Lipschitz_normalization}
\begin{aligned}
\hat W^{(\ell)}_{ij} 
    &= \text{normalization}\!\left(W^{(\ell)}_{ij},\, c_\ell\right),
    \qquad c_\ell > 0, \\
    &= \min\!\Bigg(1,\,
        \frac{c_\ell}{\sum_{k} \lvert W^{(\ell)}_{ik}\rvert}
       \Bigg)\, W^{(\ell)}_{ij}.
\end{aligned}
\end{equation}


\noindent\textbf{Parameterization and sensitivity.}
A key distinction between SNS-MLPs and prior Lipschitz MLPs \cite{liu2022learning} is how the learned layerwise constants $c_\ell$ are parameterized. Because the global Lipschitz upper bound is multiplicative,
\begin{equation}
C = \prod_{\ell=1}^L c_\ell,
\end{equation}
the parameterization of each $c_\ell$ determines how regularization pressure on $C$ is distributed across layers.

Lipschitz MLPs enforce positivity with
\begin{equation}
c_\ell=\operatorname{softplus}(\theta_c^{(\ell)}).
\end{equation}
For this parameterization,
\begin{equation}
\frac{\partial \log C_{\mathrm{sp}}}
{\partial \theta_c^{(j)}}
=
\frac{\sigma(\theta_c^{(j)})}
{\operatorname{softplus}(\theta_c^{(j)})}
=
r(\theta_c^{(j)}).
\end{equation}
The function $r(\theta)$ is strictly decreasing, so the regularizer applies larger updates to already-small layerwise constants than to large ones. Thus, under the product-penalty gradient, softplus can amplify imbalance across layers and drive small constants toward collapse. We prove that $r$ is strictly decreasing in Appendix~\ref{ap:softplus-exp-proof}. This analysis considers the penalty in isolation, so it identifies a mechanism that can lead to collapse rather than proving collapse; Section~\ref{sec: convergence} examines it in full training.

To avoid this asymmetric shrinkage, we instead use
\begin{equation}
c_\ell=\exp(\theta_c^{(\ell)}).
\end{equation}
Then
\begin{equation}
\log C_{\exp}
=
\sum_{\ell=1}^L \theta_c^{(\ell)},
\qquad
\frac{\partial \log C_{\exp}}
{\partial \theta_c^{(j)}}
=
1.
\end{equation}
The exponential parameterization therefore gives each layer equal log-space sensitivity to the global budget. This balanced geometry is the architectural change that distinguishes SNS-MLPs from the Lipschitz MLP.

\noindent\textbf{Smooth neural surrogate objective.}
We define the SNS objective as
\begin{equation}
\label{eq:kSNS_general}
\begin{aligned}
\mathcal{L}
&=
\mathcal{L}_{\text{main}}
+
\lambda\,
\mathcal{L}_{\text{SNS}} \\
&=
\mathcal{L}_{\text{main}}
+
\lambda\,
\max\!\left(\!
    1,\;
    \frac{C}{c_{\text{ub}}}
\right),
\end{aligned}
\end{equation}
where $c_{\text{ub}}$ is the corresponding sensitivity budget. Rescaling the numerator by $c_{\text{ub}}$, rather than enforcing $\max(C,c_{\text{ub}})$ directly, ensures that the regularization coefficient $\lambda$ remains independent of the scale of $c_{\text{ub}}$.

Although Eq.~\eqref{eq:kSNS_general} uses a hinge-type regularizer through the $\max$ operator, this does not materially affect training stability. The objective is piecewise smooth and is nondifferentiable only at the isolated transition point where the budget becomes active. In practice, this kink is benign in autodiff-based optimization, and we did not observe any instability or convergence degradation attributable to this regularizer.



\begin{figure}
    \centering
    \includegraphics[width=0.90\linewidth]{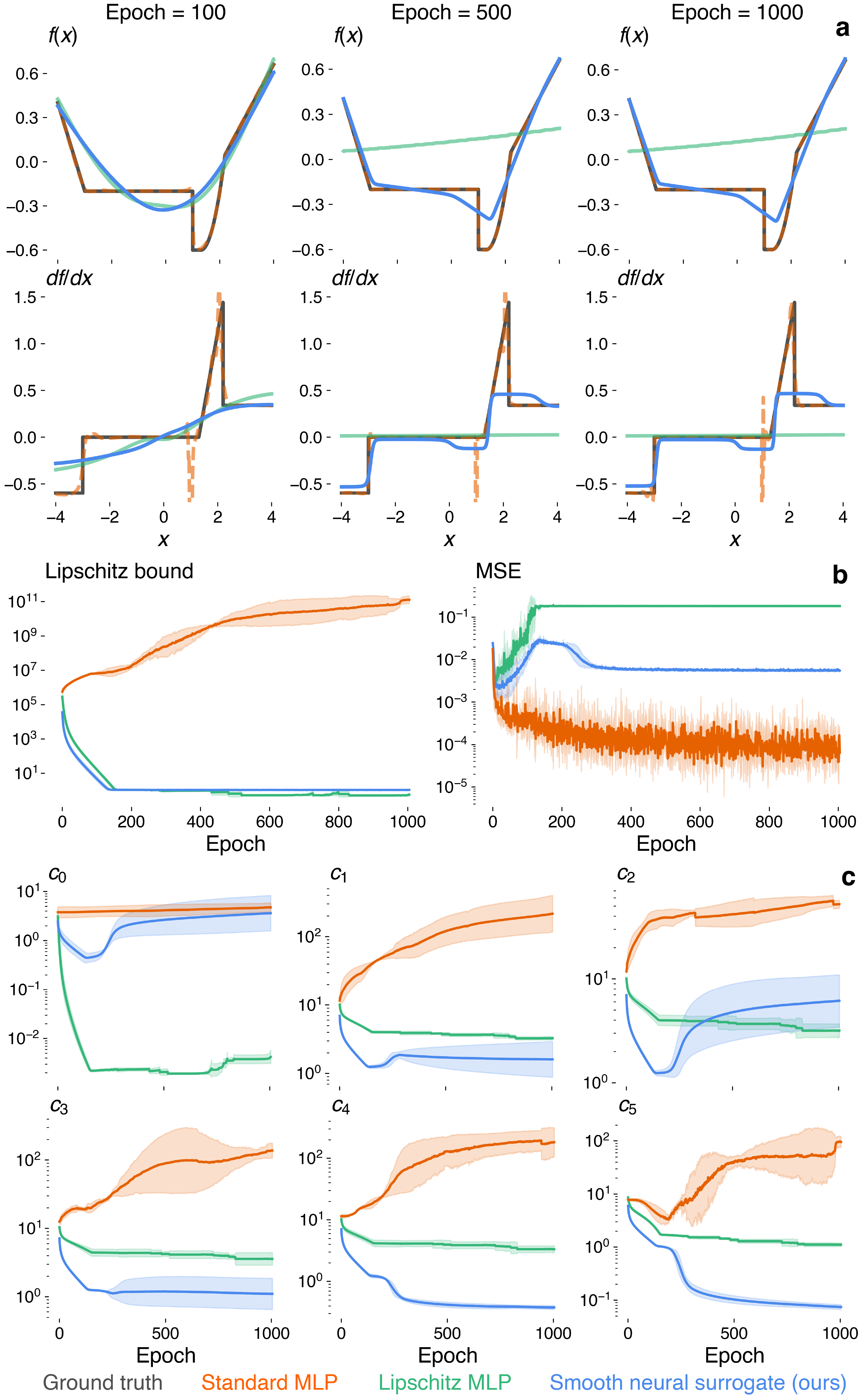}
\caption{\textbf{Smooth neural surrogates provide informative derivatives throughout training.} \textbf{a}, Evolution of model derivatives for a nonsmooth piecewise function during training. Smooth neural surrogates maintain informative derivatives suitable for optimization, whereas standard MLPs develop stiff gradients and Lipschitz MLPs \cite{liu2022learning} collapse under smoothness constraints. \textbf{b}, Evolution of Lipschitz bounds and MSE during training. \textbf{c}, Layerwise Lipschitz constants $c_\ell$ throughout training. Under the softplus parameterization, imbalance among layerwise constants grows as the Lipschitz MLP collapses, whereas the exponential parameterization avoids penalty-driven imbalance (Section~\ref{sec: smooth neural networks} and Appendix~\ref{ap:softplus-exp-proof}).}
    \label{fig:funky nonsmooth}
\end{figure}

\subsection{Experiments and Examples}
\noindent These experiments are designed to answer the following questions: (1) What do smooth neural surrogates of simple functions look like, and how do they differ from standard MLP approximations? and (2) How does the proposed parameterization of the layerwise constants $c_\ell$ affect convergence behavior during training and mitigate failure modes of Lipschitz MLPs? 

Since our method is a direct extension of Liu \emph{et al.}, who already provide extensive empirical comparisons against alternative smooth neural parameterizations and regularization strategies, we do not seek to re-establish state-of-the-art performance. Accordingly, we restrict our baselines to their Lipschitz MLP and a standard MLP. For the Lipschitz MLP baseline, we train using $\mathcal{L}_{\text{SNS}}$ to enable a controlled comparison that isolates the effect of the exponential parameterization of the layerwise Lipschitz constants. Where applicable, confidence intervals in figures are calculated from the log-transformed data. All experiments are implemented in JAX and repeated with 3 random seeds; budgets, regularization weights, learning rates, and final results are provided in Appendix~\ref{ap: sns experiments} and Table~\ref{tab:toy}.

\noindent\textbf{Shape interpolation.} A key advantage of smooth neural surrogates is their ability to learn globally smooth representations of unknown functions. We evaluate this property on a 2-D shape interpolation task, where the model must interpolate between two shapes given training samples only at $z=0$ and $z=1$. The neural networks model signed-distance functions in $(x,y)$ conditioned on the latent variable $z$, yielding $d \approx f_\theta(x,y,z)$ where $d$ is the distance to the surface of the shape. 

The standard MLP fails to produce coherent or continuous interpolation in $z$ (Fig.~\ref{fig:shape interpolation}); instead, its intermediate representation has spurious discontinuities. Although prior work reports successful interpolation with Lipschitz MLPs, this baseline collapses under a tight sensitivity budget ($c_{\text{ub}}=8$), even when the regularization weight is reduced by two orders of magnitude. In contrast, the SNS converges in both smoothness and MSE and consistently reconstructs valid intermediate shapes in a zero-shot manner. From this simple example, it is clear that smoothness can play a key role in model generalization.

\noindent\textbf{1-D nonsmooth functions.} We further study smoothness constraints on a controlled task: a 1-D piecewise function (Appendix~\ref{ap: sns experiments}) that mimics a hopping robot's cost landscape \cite{xue2025full}. In Fig.~\ref{fig:funky nonsmooth}, the standard MLP fits the data with little bias but yields uninformative gradients, and the Lipschitz MLP again collapses under the constraint. SNS-MLPs converge to stable, slightly biased representations that smooth the discontinuities and nonconvexities.



\begin{figure}
    \centering
    \includegraphics[width=0.90\linewidth]{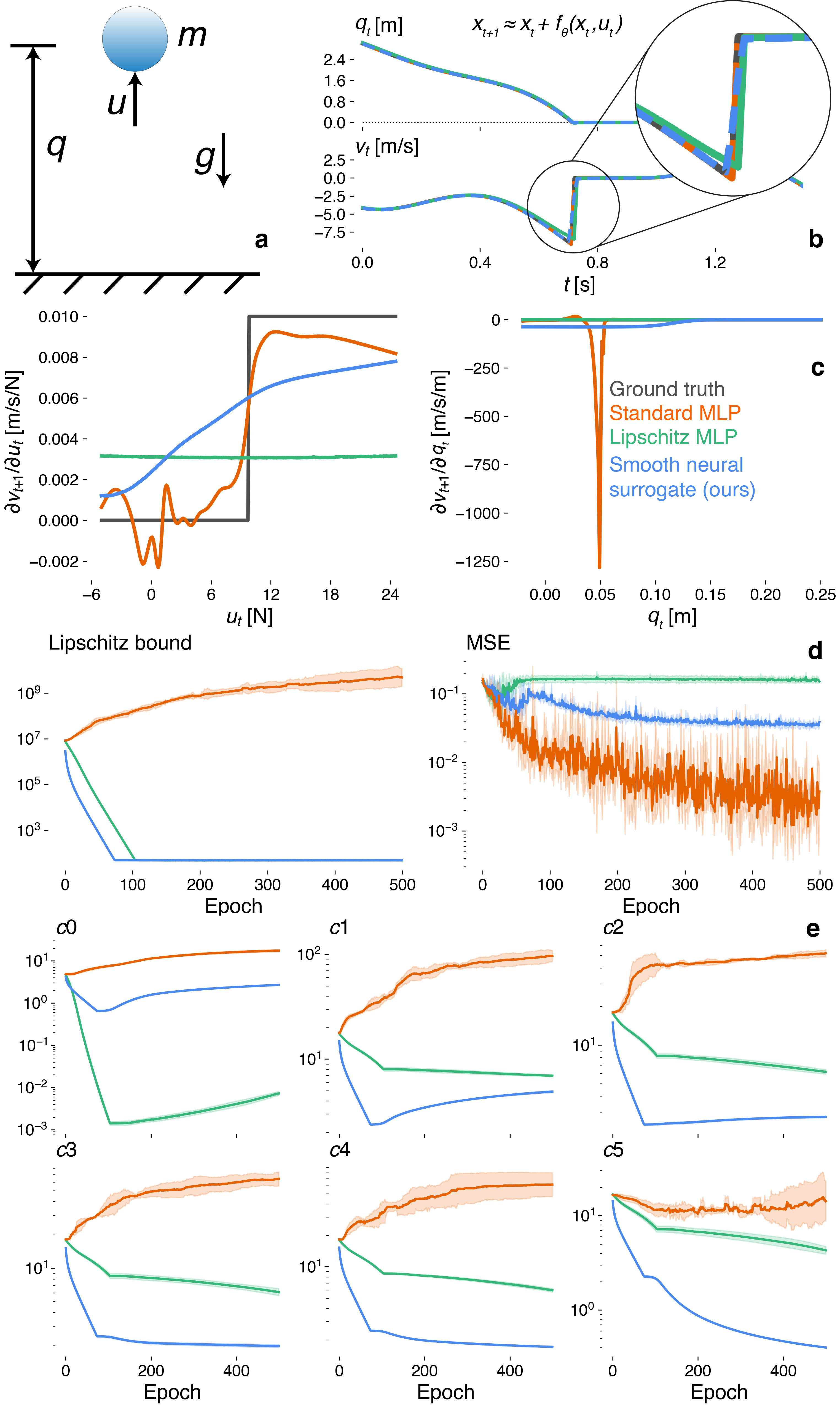}
\caption{\textbf{Smooth neural surrogates learn contact dynamics while preventing exploding gradients.} \textbf{a}, Free-body diagram of the particle--mass impact system used to generate training data. \textbf{b}, Zeroth-order predictions of three neural dynamics models. Standard MLPs closely match the ground-truth dynamics, Lipschitz MLPs collapse, and smooth neural surrogates accurately capture the dynamics while smoothing the impact transition. \textbf{c}, Derivatives of the predicted next-step velocity, $v_{t+1}$, with respect to control ($\partial u_t$) and position ($\partial q_t$). Standard MLPs exhibit stiff, erratic gradients, whereas SNS-MLPs produce smooth, attenuated, and informative gradients. \textbf{d}, Lipschitz bounds and MSE during training. \textbf{e}, Layerwise Lipschitz bounds showing collapse of the first layer in the Lipschitz MLP (Section~\ref{sec: smooth neural networks} and Appendix~\ref{ap:softplus-exp-proof}).}
    \label{fig: particle mass}
\end{figure}

\noindent\textbf{Smoothing particle–mass contact dynamics.}
Finally, we evaluate the surrogates on a physical system with stiff contact dynamics: a particle-mass model impacting the ground (Appendix~\ref{ap: sns experiments}). Networks are trained to predict next-step position and velocity $x_{t+1}\approx x_t +f_\theta(x_t, u_t)$. Standard MLPs accurately capture zeroth-order behavior but yield erratic, stiff derivatives (Fig.~\ref{fig: particle mass}). Notably, the MLP's learned derivatives can be smoother than the ground truth, consistent with prior observations \cite{parmar2021fundamental}, but are overall quite stiff and noisy. The SNS-MLP models the dynamics while maintaining stable, well-behaved derivatives.

\noindent\textbf{Layerwise Lipschitz constants.} The learned constants $c_\ell$ (Figs.~\ref{fig:shape interpolation}c, \ref{fig:funky nonsmooth}c, and \ref{fig: particle mass}e) show how each parameterization distributes the penalty. Under softplus, shrinkage concentrates in the input layer, whose constant falls below 0.03 while the other layers remain $O(1$--$10)$, and the onset of this collapse coincides with the stall in training error (Figs.~\ref{fig:shape interpolation}b, \ref{fig:funky nonsmooth}b, and \ref{fig: particle mass}d). Under the exponential parameterization, the constants shrink together while the penalty is active, and the smallest remains an order of magnitude or more above its softplus counterpart (Table~\ref{tab:toy}). This pattern matches the gradient imbalance analyzed above, and Section~\ref{sec: convergence} shows a similar effect in the quadruped dynamics.

\section{Learning Generalizable Legged MPC and State Estimation}\label{sec: dynamics and estimation}
\noindent In the previous section, we showed that standard neural networks are often unreliable as smooth surrogates for nonsmooth functions: they both inherit the nonsmoothness from the training data and introduce additional spurious nonsmoothness, particularly when data are sparse in regions of the input space. To address these limitations, we introduced the smooth neural surrogate MLP, which mitigates such effects through explicit control of global Lipschitz continuity.

In this section, we detail our methodology for scaling these lightweight architectures to quadruped robots and model predictive control under domain randomization and partial observability. The full training procedure is summarized in Algorithm~\ref{alg:full-training}, with implementation details provided in the sections that follow and in Appendix~\ref{ap: quadruped experiments}.

\subsection{Preliminaries}
\noindent \textbf{Learned MPC.} Given states $x_t$, and actions $u_t$, a typical formulation \cite{lutter2021learning, chua2018deep} for learned dynamics is
\begin{equation}
\label{eq:dynamics_net_general}
\begin{aligned}
x_{t+1}
= x_t + d\hat{\mu}_\theta(x_t, u_t)dt + \varepsilon,
\qquad
\varepsilon \sim P,
\end{aligned}
\end{equation}
where $P$ denotes an unspecified residual or error distribution. It is common to train the dynamics with one-step or multi-step Gaussian losses. It is also standard for the dynamics to depend on a history of states and actions ($x_{t-H:t}, u_{t-H:t}$) to improve accuracy or compensate for partial observability \cite{xu2025neural, moore2025automated}. As the dynamics are learned, we solve a variation of the following standard optimal control problem (OCP):
\begin{equation}
\begin{aligned}
\min_{x_{0:T}, \,u_{0:T-1}} \quad & \sum_{t=0}^T\ell\big(x_t, u_t\big)  \\
\text{s.t.} \quad  
 &x_{t+1}
- x_t - d\hat{\mu}_\theta(x_t, u_t)dt =0\\
   \quad  
 &x_0-x^{\text{est.}}_0  = 0 \\
  \quad  
 &g_i\big(x_t, u_t\big)  \leq 0.
\end{aligned}\label{eq: rollout OCP}
\end{equation}
The OCP is recast in single-shooting form using differentiable rollouts. We compare two methods for solving \eqref{eq: rollout OCP}: a state-of-the-art sampling-based method, DIAL-MPC \cite{xue2025full}, and our generalized Gauss--Newton method (GGN-MPC). The optimization is vectorized across environments using \texttt{vmap}$(\cdot)$.

\noindent \textbf{Learned state estimation.} Since legged robots operate under partial observability, we need state estimation in order to use the learned dynamics for MPC. Classical estimators for legged robots, such as Kalman or particle-filter variants, often rely on accurate contact information, prior gait knowledge, or rigid-body assumptions, and they can become unreliable on varied terrain \cite{kim2021legged, hartley2020contact, ji2022concurrent}. Although learned state estimation is increasingly common, most concurrent learning for estimation and control in legged robots appears in model-free RL \cite{ji2022concurrent, kim2025learning}.  Other approaches to neural state estimation take inspiration directly from classical estimators and exploit model predictions \cite{revach2022kalmannet}. 


\begin{figure}
    \centering
    \includegraphics[width=\linewidth]{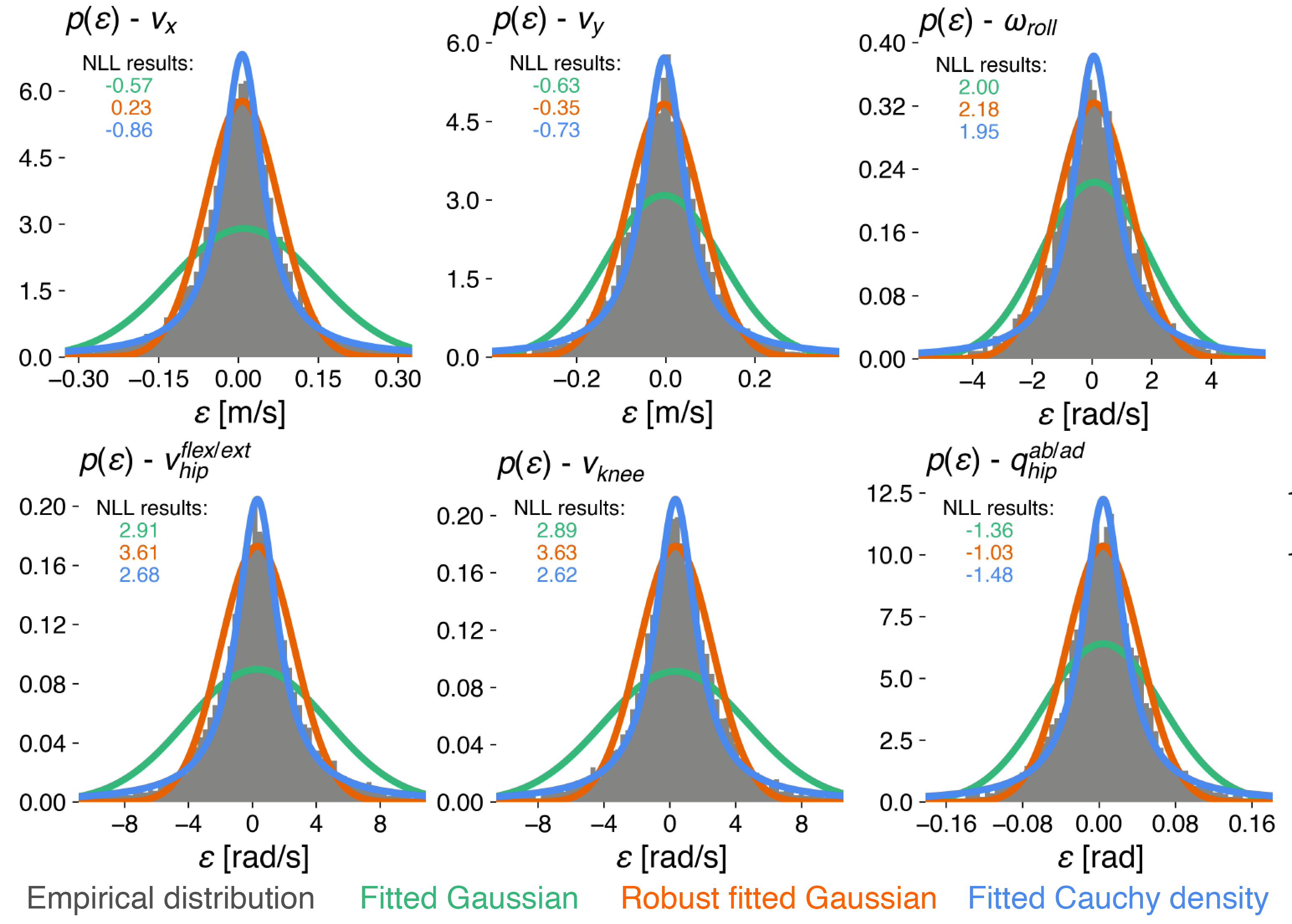}
    \caption{\textbf{Errors for learned quadruped dynamics follow a heavy-tailed Cauchy distribution.} Empirical one-step prediction residuals $\varepsilon$ across representative quadruped state variables. Residuals computed from the replay buffer align more closely with a Cauchy density (lower NLL) than with a Gaussian, even when the latter is fit using robust estimators. This suggests that Gaussian losses, such as MSE, are poorly suited for learning legged robot dynamics. Robust Gaussian estimates exploit the identity that, for a true Gaussian, the mean equals the median and $\mathrm{STD}\approx1.4826\,\mathrm{MAD}$.}
    \label{fig: empirical distributions}
\end{figure}


\noindent \textbf{Heavy-tailed maximum likelihood estimation.} Gaussian error models, implemented through common objectives such as MSE, are a reasonable default. In our setting, however, we observe that the Cauchy distribution, which is also bell-shaped but heavy-tailed, better reflects the empirical residuals (Fig.~\ref{fig: empirical distributions}). The Cauchy distribution is the canonical example of a ``pathological'' distribution. Unlike the Gaussian, the Cauchy distribution has no finite mean or variance and samples from this distribution over time are described as ``jumps'' or ``impulses'', not unlike transitions in hybrid dynamics (Fig.~\ref{fig: high-level idea}). 

If we assume $\varepsilon\sim \mathcal{C}(x; \mu, \Sigma)$, we want to minimize any loss proportional to the \emph{Cauchy negative log-likelihood} during training. In the multi-dimensional case:
\begin{equation}\label{eq: CNLL}
\begin{aligned}
\!-\log \mathcal{C}(x;\mu,\Sigma) &= \tfrac{n+1}{2}\log(1+(x-\mu)^\top  \Sigma^{-1}(x-\mu)) \!\\&+\tfrac{1}{2}\log |\Sigma|
 + \text{const.},
\end{aligned}
\end{equation}
which follows from a special case of the Student's t-density. The parameter $\mu$ defines the \emph{location}, while $\Sigma$ acts as a \emph{dispersion} term controlling scale along each direction. In the isotropic case $\Sigma = \sigma^2 I$, the location coincides with the median, and the dispersion parameter $\sigma$ is proportional to the median absolute deviation (MAD) from the median. 

Variants of the Cauchy likelihood automatically rescale gradients. For small errors, both Cauchy and Gaussian losses behave quadratically. For large errors, the Cauchy gradient saturates, preventing instability and reducing the impact of large $\varepsilon$. For cases where the impulse-like dynamics are easy to predict, like our particle-mass example, the two losses will perform similarly due to this comparability under small $\varepsilon$.
\begin{figure}
    \centering
    \includegraphics[width=\linewidth]{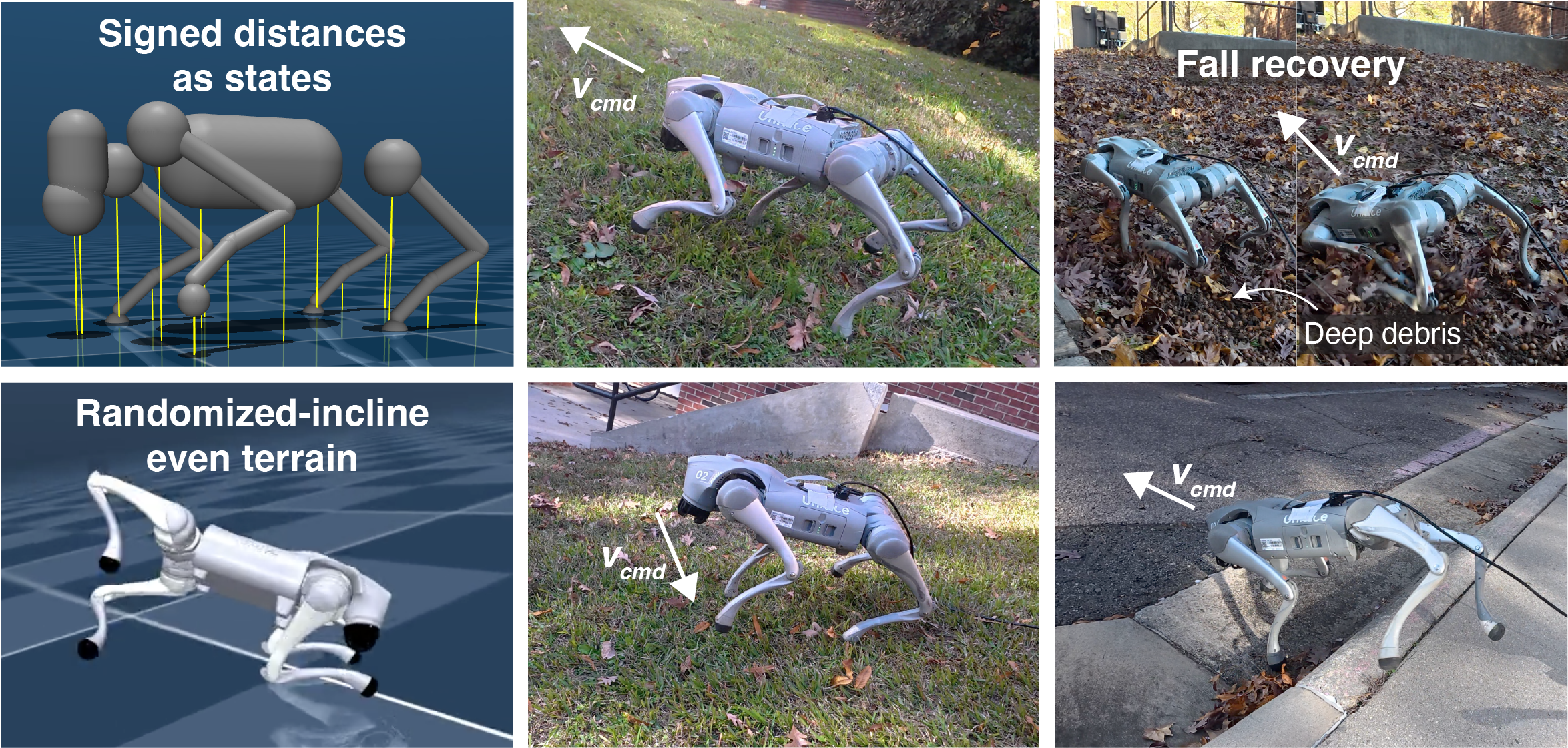}
    \caption{\textbf{A strategy for whole-body and terrain-aware planning with learned MPC.} By augmenting the states in our neural dynamics model with signed distances from the robot to the terrain, the controller can plan using both classical generalized coordinates and spatial relationships between the body and environment. Training with randomized terrain inclines makes the state estimates and plans robust to uneven ground, varying slopes, and stepped terrain in the real world. Even on sloped, slippery ground with deep debris, the model demonstrates precise awareness of collision geometry and contact dynamics, allowing it to regain footing after a fall.}
    \label{fig: whole-body}
\end{figure}
\subsection{Partially Observable Dynamics and State Estimation}

\noindent\textbf{Learned dynamics.}
We categorize all quantities into measurements $y_t$ (e.g., linear acceleration),
state variables $x_t$ used in control (e.g., velocity), and unmeasured
environmental factors $z_t$ (e.g., friction). The effect of $z_t$ is captured implicitly
through time-delay embedding \cite{abarbanel2012analysis}. Let
\begin{equation}
X_t := x_{t-H:t},
\qquad
U_t := u_{t-H:t},
\end{equation}
denote fixed-length state and action histories. Our
dynamics modify \eqref{eq:dynamics_net_general} as
\begin{equation}\label{eq:dyn_chi}
\begin{aligned}
x_{t+1}
&=
x_t
+
d\hat{\mu}_\theta(X_t, U_t)\, dt
+
\varepsilon,
\qquad
\varepsilon \sim \mathcal{C}(0,\hat{\Sigma}) \\
&\approx \hat\mu_\theta(X_t, U_t)
\end{aligned}
\end{equation}

\noindent\textbf{Learned estimator (predictor--corrector).}
Because the dynamics depend on histories, the estimator must maintain a
history estimate defined as:
\begin{equation}
\bar X_{t|t-1} := \bar x_{t-H:t|t-1},
\qquad
\bar X _{t|t} := \bar x_{t-H:t|t},
\end{equation}
where $\bar X_{t|t-1}$ is the prior (predicted) history and $\bar X_{t|t}$ is the
posterior (corrected) history.

\noindent \textit{Prediction step.} The estimator rolls its previous posterior one step forward using the learned
dynamics:
\begin{equation}
\bar X_{t|t-1}
=
\texttt{roll}\!\left(
\bar X_{t-1|t-1},
\hat{\mu}_\theta(\bar X_{t-1|t-1}, U_{t-1})
\right),
\label{eq:X_bar_predict}
\end{equation}
where $\texttt{roll}(a, b)$ discards the oldest element of $a$ and appends $b$. Measurements are predicted via the prior and a typical measurement model $h(\cdot)$, then used to calculate the innovation:
\begin{equation}
\hat{y}_t = h(\bar X_{t|t-1}),
\qquad
\nu_t = y_t - \hat{y}_t.
\label{eq:innov}
\end{equation}
Then, the gradient of the squared innovation norm provides sensitivities for the neural estimator:
\begin{equation}
g_{\nu,t}
=
\begin{bmatrix}
\frac{\partial(\nu_t^\top \nu_t)}{\partial \bar X_{t-1|t-1}} &
\frac{\partial(\nu_t^\top \nu_t)}{\partial U_{t-1}}
\end{bmatrix}^\top.
\label{eq:innov_grad_exact}
\end{equation}

\noindent \textit{Correction step.} The estimator applies a correction to the prior using all available information:
\begin{equation}\label{eq:X_bar_correct}
\begin{aligned}
X_t
&=
\bar X_{t|t-1}
\!+
d\bar\mu_\zeta\!\left(
\bar X_{t|t-1},
U_t,
Y_t,
\nu_t,
g_{\nu, t}
\right)\!
+
\varepsilon,\,\,
\varepsilon \sim \mathcal{C}(0,\hat{\Sigma}) \\
&\approx\bar\mu_\zeta\!\left(
\bar X_{t|t-1},
U_t,
Y_t,
\nu_t,
g_{\nu, t}
\right).
\end{aligned}
\end{equation}
where $\{Y_t: y_{t-H:t} \cap X_t = \emptyset\}$ are the measurement histories that do not overlap with the state histories. This would include, for example, base acceleration. For directly measurable state components (e.g., joint angles and velocities),
we fill the corresponding entries of $\bar X_{t|t}$ with their noise corrupted sensor
readings rather than a learned correction. In other words, the estimator is only trained to estimate unmeasurable or ``privileged'' states while noisy measurements/states are used without correction.

\noindent \textbf{Loss functions and concurrent training.} We train the dynamics model and estimator \emph{concurrently}, with each module receiving its own loss while also influencing the other’s inputs during rollout. This is not unlike concurrent methods in model-free RL \cite{ji2022concurrent}. Algorithms~\ref{alg:dynamics step} and~\ref{alg:estimator step} detail how the losses are computed during training. The dynamics model is trained with
\begin{equation}
\begin{aligned}
        \mathcal{L}_{\text{dynamics}} = \lambda_0 \mathcal{L}_{\text{step}} 
        + \lambda_1 \mathcal{L}_{\text{rollout}} 
        + \lambda_2 \mathcal{L}_{\text{corrupt}}
        + \lambda_3 \mathcal{L}_{\text{SNS}} .
\end{aligned}\label{eq: dynamics loss}
\end{equation}
For robust optimization under heavy-tailed residuals we use the mean Cauchy error (MCE)
\begin{equation}\label{eq: MCE}
    \mathcal{L}_{\text{MCE}} \propto -\log \mathcal{C}(x_{t+1}; \hat\mu_\theta, \hat\Sigma_\mathcal{D}),
\end{equation}
with homoscedastic dispersion $\hat{\Sigma}_\mathcal{D}=\hat{\sigma}^2 I$ estimated via the replay buffer and the MAD. The same MAD and median values are used for fixed ``normalization'' and ``denormalization'' layers for the input and output. In the homoscedastic case, \eqref{eq: MCE} corresponds to dropping the $\log|\Sigma|$ term in \eqref{eq: CNLL} and averaging across the state and batch dimensions. 
\begin{figure*}[t]
    \centering
    \includegraphics[width=\linewidth]{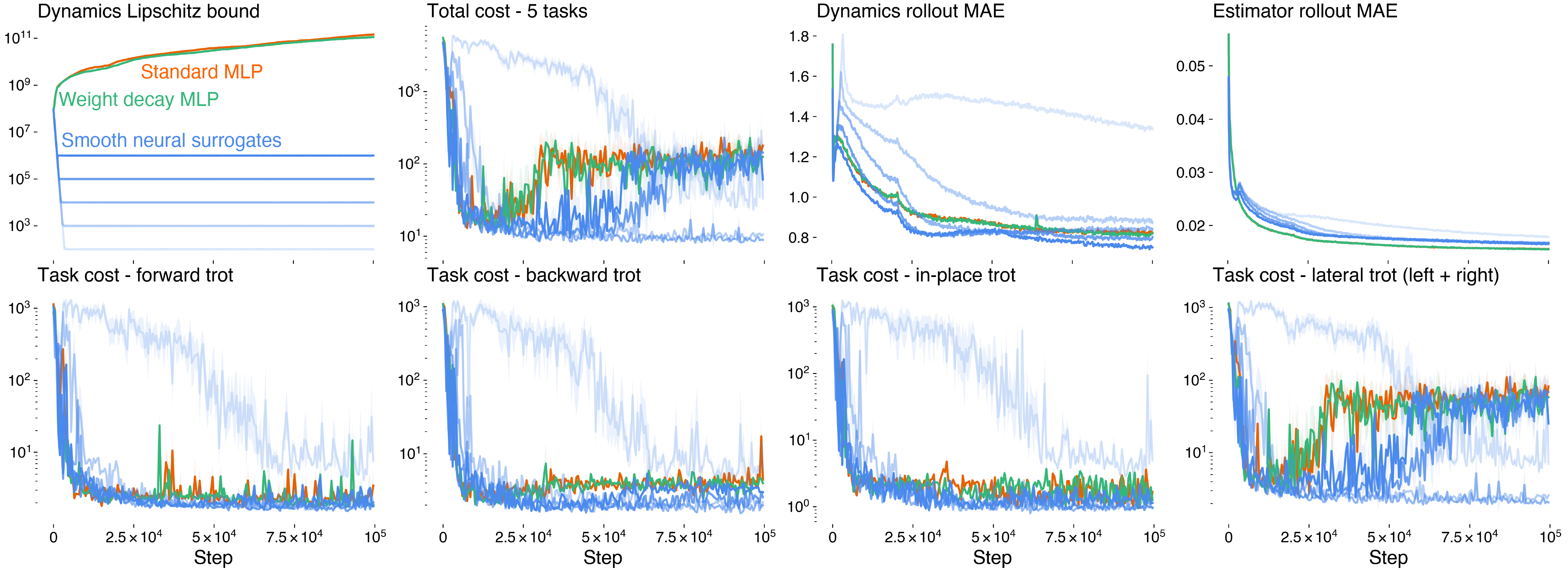}
\caption{\textbf{Smooth neural surrogates improve learned MPC.} SNS dynamics with varying Lipschitz constraints are compared to standard and weight-decay MLPs. Moderate smoothness gives the best control tradeoff: overly tight bounds slow learning, while loose or unconstrained models produce stiff dynamics that degrade planning despite competitive prediction error.}
\label{fig: smooth vs others}
\end{figure*}
\begin{figure}
\centering
\includegraphics[width=\linewidth]{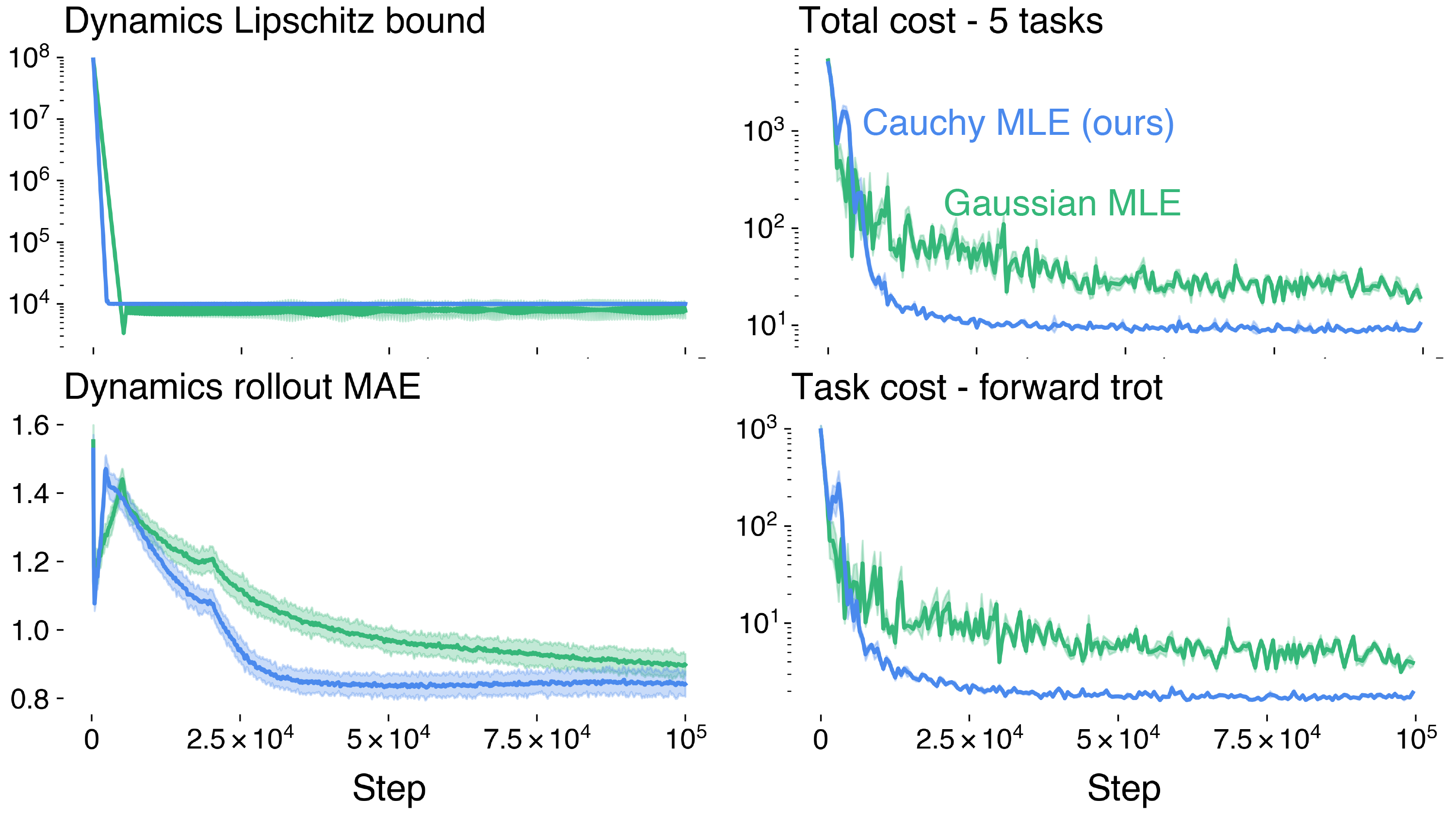}
\caption{\textbf{Learning with heavy-tailed maximum likelihood estimation improves training stability, motion planning, and model accuracy.} Gaussian MLE, common in model-based RL, fails to capture the error distribution we observe while training stiff legged-robot dynamics. In contrast, heavy-tailed Cauchy MLE yields more stable convergence, higher prediction accuracy, and lower overall control cost. With Cauchy likelihoods we observe \textbf{$\approx$5$\times$ lower cost} at 25,000 training steps, shrinking to \textbf{$\approx$2-3$\times$ lower cost} at 100,000 steps.}
\label{fig: mse vs cauchy}
\end{figure}

\begin{figure}[t]
    \centering
    \includegraphics[width=\linewidth]{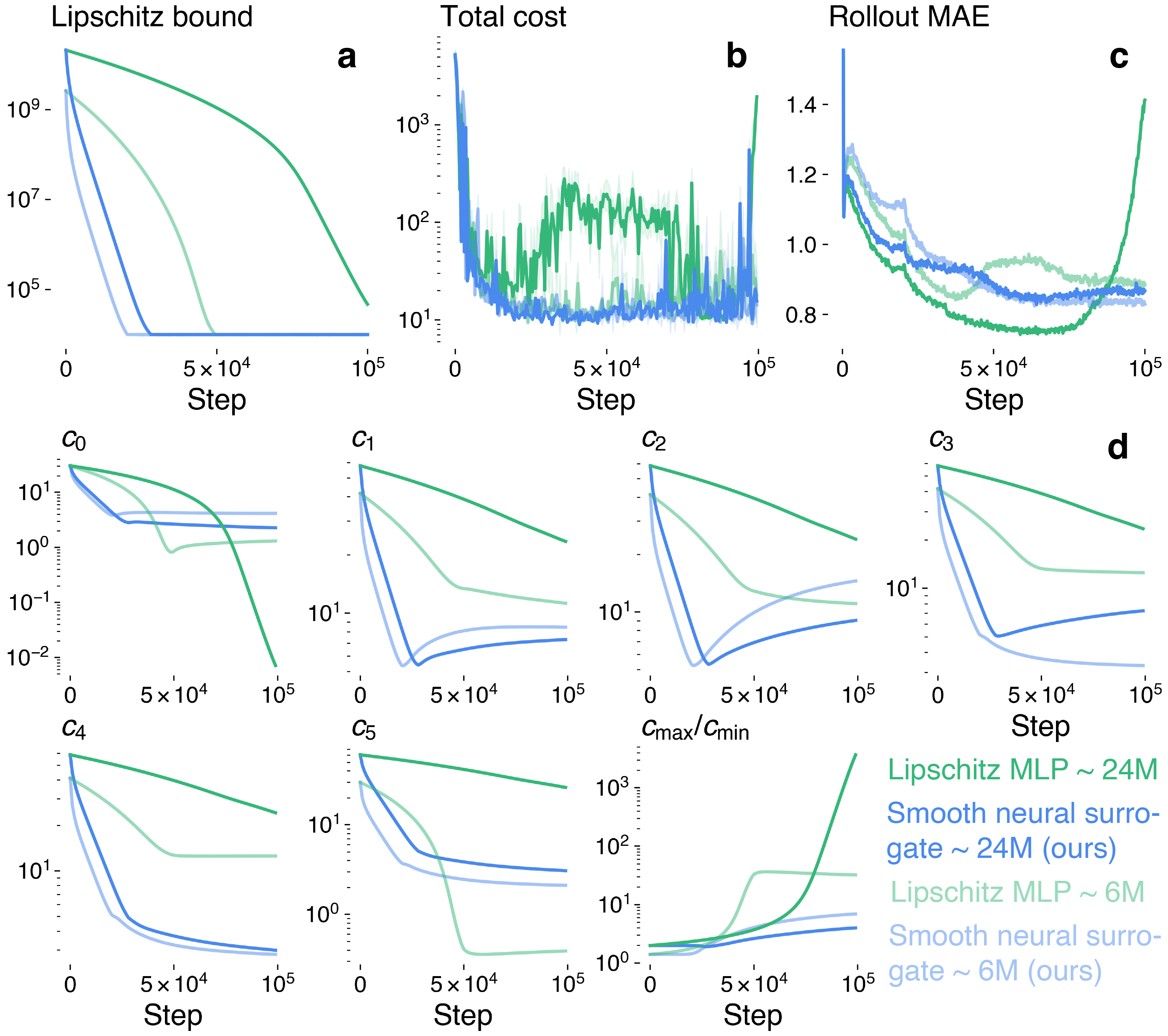}
\caption{\textbf{Isolating the parameterization in learned legged MPC.} SNS and Lipschitz MLP \cite{liu2022learning} dynamics at two widths, differing only in the parameterization of the layerwise constants $c_\ell$ ($c_{\text{ub}}=10^4$). \textbf{a}, Lipschitz upper bounds. \textbf{b}, Closed-loop cost summed over five trotting tasks. \textbf{c}, Dynamics rollout MAE. \textbf{d}, Layerwise constants and the ratio $c_{\max}/c_{\min}$. The wide Lipschitz MLP stays far above its budget for most of training and fails once its input-layer constant collapses.}
\label{fig: lip vs sns mpc}
\end{figure}

The single-step loss along a trajectory of length $T$ is
\begin{equation}\label{eq: L_step}
    \mathcal{L}_{\text{step}} = \tfrac1T\sum^{T}_{t=0}
    \mathcal{L}_{\text{MCE}}(x_{t+1}, \hat\mu_\theta, \hat\Sigma_\mathcal{D}),
\end{equation}
while the multi-step rollout loss is computed autoregressively with discount $\gamma$:
\begin{equation}\label{eq: L_rollout}
    \mathcal{L}_{\text{rollout}} = 
    \tfrac1{T-1}\sum^{T}_{t=1}
    \mathcal{L}_{\text{MCE}}(x_{t+1}, \hat\mu_{\theta,t+1}, \hat\Sigma_{\mathcal{D}, \gamma^t}).
\end{equation}
This prevents long-horizon drift \cite{moore2025automated, lutter2021learning}. The corrupted-input loss uses estimator outputs as dynamics inputs:
\begin{equation}\label{eq: L_corrupt}
    \mathcal{L}_{\text{corrupt}}
    = \tfrac1{T-1}\sum^{T}_{t=1}
    \mathcal{L}_{\text{MCE}}(
    x_{t+1},\,
    \hat\mu_\theta(\texttt{sg}(\bar X_{t|t}), \cdot),
    \hat\Sigma_{\mathcal{D}})
\end{equation}
to ensure robustness to poor estimates. The estimates are treated as plain data via the stop gradient operator, $\texttt{sg}(\cdot)$, to avoid training instability. The algorithm for a full training step of the dynamics model is given in Algorithm~\ref{alg:dynamics step}.

The estimator is trained to converge over time using multi-step rollouts starting from noise-perturbed histories:
\begin{equation}
\begin{aligned}
        \mathcal{L}_{\text{estimator}} &=
        \tfrac1{H(T-1)}\sum^{T}_{t=1}\sum^{H}_{h=0}
        \mathcal{L}_{\text{MCE}}(
        x_{t-h:t},\,
        \bar\mu_\zeta(\bar X_{t\mid t-1}, \cdot),\,
        \hat\Sigma_{\mathcal{D}})
        \\
        &\quad + \lambda_4 \mathcal{L}_{\text{SNS}} .
\end{aligned}\label{eq: estimator loss}
\end{equation}
As before, we detach the estimates to prevent exploding gradients during the rollout. The algorithm for a full training step of the estimator model is given in Algorithm~\ref{alg:estimator step}.

\subsection{Generalized Gauss--Newton MPC}
\noindent As our primary gradient-based solver for the OCP, we use a generalized Gauss--Newton sequential quadratic programming method~\cite{schraudolph2002fast, grandia2023perceptive, messerer2021survey}. We call the approach \emph{gray-box}: beyond differentiability and convex costs, the optimizer is agnostic to the internal structure of the learned dynamics model. This contrasts with Gauss--Newton trajectory optimizers such as iLQG~\cite{todorov2005generalized}, which impose specific temporal structure on the dynamics, actions, and costs. Our method is inspired by iLQG but does not require these structural constraints, and we accelerate the updates on the GPU.

We parameterize the action (control) sequence $u_{0:T-1}$ using a set of $k$ knots $u_\kappa$:
\begin{equation}
    u_t = \texttt{spline}(u_\kappa)(t), \quad u_\kappa\in \mathbb{R}^{n\cdot k},\quad k\leq T.
\end{equation}
This parameterization is common in gradient- and sampling-based methods, as it enforces continuity and physical plausibility while reducing the number of decision variables.

Constraints are handled using an augmented objective following \cite{grandia2023perceptive, kim2025learning}:
\begin{equation}
\begin{aligned}
\min_{u_\kappa} \quad & \ell(u_\kappa) + \sum^{G}_{i=0}\beta_i\log_{\delta}(g_i(u_\kappa)).
\end{aligned}\label{eq:obj_relaxed_barrier}
\end{equation}
Here, $\log_\delta(\cdot)$ denotes a relaxed logarithmic barrier:
\begin{equation}
\log_\delta(g_i)=
\begin{cases}
-\log(-g_i), & g_i<-\delta_i, \\[6pt]
-\log\delta_i+\tfrac{1}{2}\left(\tfrac{g_i+2\delta_i}{\delta_i}\right)^2-\tfrac{1}{2}, & g_i\geq-\delta_i,
\end{cases}
\end{equation}
The relaxation parameters $\delta_i$ and $\beta_i$ regulate the strength of constraint enforcement. As $\delta_i \to 0$ and $\beta_i \to 0$, feasible solutions incur vanishing barrier cost, whereas infeasible solutions diverge. In practice, we keep these parameters fixed.

\noindent\textbf{Backward pass / quadratic approximation.}  
In the backward pass, we compute a local descent direction for the spline knots. Let \eqref{eq:obj_relaxed_barrier} be written as a sum over residuals $\varepsilon_i$:
\begin{equation}
\mathcal{J}(\varepsilon(u_\kappa))
= \sum_{i=0}^{R} \ell_i\big(\varepsilon_i(u_\kappa)\big).
\label{eq: residual cost}
\end{equation}
Here, “residuals” refers to the per-term objective errors in \eqref{eq: residual cost} and should not be conflated with the learned dynamics or state estimation residuals discussed above. We seek an update direction $\delta u_\kappa$ at point $u_\kappa^-$ by solving the subproblem:
\begin{equation}
\min_{\delta u_\kappa}\; 
\tfrac{1}{2}\,\delta u_\kappa^\top H\,\delta u_\kappa 
\;+\; q^\top \delta u_\kappa.
\label{eq:qp}
\end{equation}
where the gradient is
\begin{equation}
q 
= \nabla_{u_\kappa}\mathcal{J}(\varepsilon(u_\kappa^-))
= 
\left(\frac{\partial \varepsilon}{\partial u_\kappa}\right)^{\!\!\top}
\nabla_\varepsilon \mathcal{J}(\varepsilon(u_\kappa^-)).
\end{equation}
and the GGN Hessian is approximated by
\begin{equation}
\begin{aligned}
H 
&\approx 
\nabla^{2}_{u_\kappa}\mathcal{J}(\varepsilon(u_\kappa^-)) \\
&=
\left(\frac{\partial \varepsilon}{\partial u_\kappa}\right)^{\!\!\top}
\nabla_\varepsilon^{2}\mathcal{J}(\varepsilon(u_\kappa^-))
\left(\frac{\partial \varepsilon}{\partial u_\kappa}\right).
\end{aligned}
\end{equation}

If each $\ell_i(\varepsilon)$ is convex in the residual, then its residual-space Hessian is positive semi-definite (PSD): $\nabla_\varepsilon^{2} \mathcal{J}(\varepsilon) \succeq 0$. By congruence, $H$ is also PSD. Then, the linear system
\begin{equation}
H\,\delta u_\kappa = -q
\end{equation}
is solved efficiently using Cholesky factorization. The rollout Jacobian $\partial \varepsilon/\partial u_\kappa$ includes derivatives through the learned dynamics and is computed using automatic differentiation.



\noindent\textbf{Forward pass / greedy parallel line search.} Line search stabilizes updates when the full Gauss--Newton step is unreliable due to nonlinearities in the rollout or cost landscape. In the forward pass, the control knots are updated along the descent direction with step size $\alpha \in [0,1]$:
\begin{equation}
u^{+}_\kappa = u^{-}_\kappa + \alpha\,\delta u_\kappa.
\end{equation}
The step size is selected by solving
\begin{equation}
\alpha^\ast
=
\arg\min_{0 \le \alpha \le 1}
\mathcal{J}\!\left(\varepsilon(u^{-}_\kappa + \alpha\,\delta u_\kappa)\right),
\end{equation}
with the full nonlinear rollout and cost. We approximate this 1D subproblem using a \emph{greedy parallel line search}: $R$ candidate step sizes $\boldsymbol{\alpha}$ are sampled uniformly in $[0,1]$, and the corresponding rollout trajectories are evaluated in parallel on GPU:
\begin{equation}
\hat{\mathbf{x}}_{0:T} 
=
\texttt{vmap}\big(\texttt{rollout}_\theta(u^{-}_\kappa + \alpha\,\delta u_\kappa)\big)(\boldsymbol{\alpha}).
\end{equation}
where $\hat{\mathbf{x}}_{0:T}\in\mathbb{R}^{R\times T \times n}$ is the batch of candidate trajectories.
\subsection{Offline Training for Online Behavior Synthesis}
\noindent \textbf{States, actions, and measurements.} We select states to support downstream planning while remaining compatible with the Lipschitz constraints of our smooth neural surrogate models. The augmented state (Table~\ref{tab:states}) includes standard proprioceptive terms together with a 6D orientation representation $r_\mathcal{W}^{6D}$ \cite{zhou2019continuity} and the signed distances $\phi$ from each body geometry to the terrain (visualized in Fig.~\ref{fig: whole-body}). These distances allow the robot to infer proximity to contact across varying terrain profiles for better awareness and can be used for collision avoidance or gait tracking costs and constraints.

\begin{table}[t]
\caption{Augmented State Space, Actions, and Measurements}
\label{tab:states}
\vspace{-5pt}
\centering
\begin{tabular}{ll}
\hline
\noalign{\vskip 1pt}
\textbf{Quantity} & \textbf{Description} \\
\hline
\noalign{\vskip 1pt}
$z$ [m] & Base height above terrain \\
$r_\mathcal{W}^{6D}$ [--]& Continuous 6D base orientation representation \cite{zhou2019continuity} \\
$q_J$ [rad] & Joint positions (12 DoF) \\
$v_J$ [rad/s]& Joint velocities \\
$v_\mathcal{B}$ [m/s]& Base linear velocity (body frame) \\
$\omega_\mathcal{B}$ [rad/s]& Base angular velocity (body frame) \\
$\phi$ [m]& Signed distances (23 collision geoms) \\
 & --------------------------------- \\

$y$ & Measurements $[q_J,\, v_J,\, r_\mathcal{W}^{\mathrm{6D}},\,
        \omega_\mathcal{B},\, \dot v_\mathcal{B}]$ \\

$\hat{y}$ & Predicted measurements $[\hat q_J,\, \hat v_J,\, \hat r_\mathcal{W}^{\mathrm{6D}},\,
              \hat\omega_\mathcal{B},\, \Delta\hat v_\mathcal{B}]$ \\
$\dot v_\mathcal{B}$ [m/s$^2$]& Base linear acceleration (body frame) \\
$\Delta\hat v_\mathcal{B}$ [m/s$^2$]& Predicted acceleration $(\hat v_{\mathcal{B},t} - \hat v_{\mathcal{B},t-1})/dt$ \\
$u$ [rad]& Desired joint positions $q_J^d$ \\ [1pt]
\hline
\end{tabular}
\end{table}




\noindent \textbf{Domain randomization.} Signed-distance states become richer when the terrain is varied. We randomize the ground-plane slope by applying a uniformly sampled 3D rotation, creating broad terrain variation without explicit curricula. Combined with the signed-distance representation, this serves as our primary mechanism for encouraging whole-body and terrain awareness (Fig.~\ref{fig: whole-body}). Although richer terrain models might further improve versatility and sim-to-real transfer, this simple scheme proved sufficient for our purposes. Other robot parameters and initial states are varied during training with full domain randomization; details are provided in Appendix~\ref{ap: quadruped experiments}.

\noindent \textbf{Control costs and constraints.} During training, we intentionally use a simple objective—no gait tracking, no tuned coefficients, and no behavior-specific shaping—to avoid implicit biases toward particular gaits. The reference commands for the base height, orientation, and velocity are randomized to a fixed value that is reset halfway through each episode in each environment (Appendix~\ref{ap: quadruped experiments}). At deployment, locomotion behaviors (trot, bound, pace, gallop, rear, tripod, and transitions) are obtained solely by adding cost terms, imposing constraints, or adjusting weight magnitudes.


\section{Experimental Results}\label{sec: results}

\noindent First, we restate our key motivating questions:
\begin{itemize} 
\item Can smooth neural dynamics and heavy-tailed likelihoods improve training stability and enable reliable gradient-based neural MPC through contact, and what role does each component play in the resulting performance? 
\item Although sampling-based MPC is often favored under learned neural dynamics, do our design choices also benefit sampling-based methods, and what advantages remain for gradient-based MPC once smoothness is enforced? 
\item Can learned MPC and state estimation for legged robots combine the versatility afforded by domain randomization with the task-level flexibility of MPC? 
\end{itemize}

We evaluate these questions through online quadruped behavior synthesis or zero-shot generalization, which we define as executing MPC tasks specified by the cost function, constraints, or environment that lie outside the training distribution. In controlled experiments, we test these motivating questions and our hypothesis that nonsmoothness is a key contributor to poor control performance. We perform targeted ablations to isolate the effects of each design choice and compare them against baselines that ignore the role of smoothness. We also present detailed experiments comparing sampling-based and gradient-based control in the context of learned MPC. Finally, we validate our conclusions through deployment on real hardware. More detailed discussions on the implementations, hyperparameters, and deployment are provided in Appendix~\ref{ap: quadruped experiments}. 


\subsection{Convergence Behavior and Basic Control}\label{sec: convergence}
\noindent We first examine how smooth neural surrogates and heavy-tailed likelihoods influence the convergence behavior, training stability, and basic trotting locomotion control. For these experiments, we simply use our gradient-based GGN-MPC (with state estimation) and evaluate the closed-loop cumulative cost over a 5 second episode, averaged over at least 10 seeds (Appendix~\ref{ap: quadruped experiments}). As before, confidence intervals within figures are calculated from the log-transformed data. In this subsection, each training setup is identical except for the independent variable of interest and the learning rate, which is tuned to the best of our ability for each method (except in the parameterization comparison, where it is shared). Smooth neural surrogates and Cauchy likelihoods are used unless otherwise stated.

\noindent \textbf{Lipschitz bounds.} Fig. \ref{fig: smooth vs others} shows how the Lipschitz bounds of MLPs influence training and control. We see that the Lipschitz bounds of standard and weight-decay MLPs grow rapidly during training, and these models perform poorly in control, especially as training continues. SNS-MLPs remain within their sensitivity budgets and exhibit a Pareto front between smoothness and control: overly smooth models learn slowly, loosely bounded models behave like standard MLPs, and moderate bounds work best for planning. 

Interestingly, control performance for lateral trotting degrades sharply, which suggests that early stopping could provide some of the benefits of smoothing. Lateral locomotion is inherently less stable, and this sensitivity likely amplifies the effects of stiff gradients, local overfitting, or nonsmooth behavior. We examine these generalization failures in greater detail in the following subsection. Crucially though, meaningful improvements in closed-loop control beyond 50,000 training steps are only observed for SNS-MLPs with $c_{\text{ub}} \leq 10^4$.

\begin{figure*}[t]
    \centering
    \adjustbox{trim=0 0 0 {.11\height},clip}{\includegraphics[width=\linewidth]{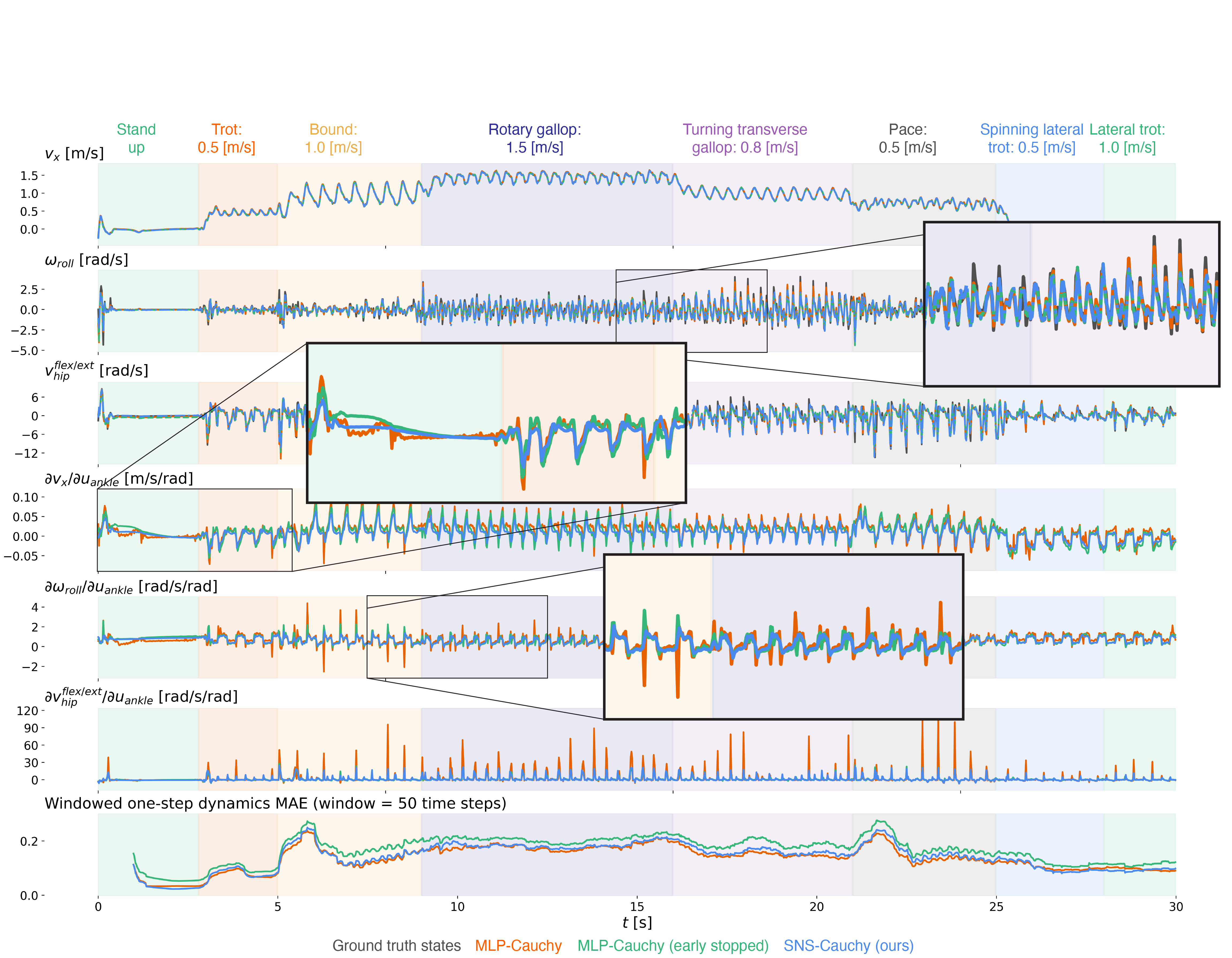}} 
\caption{\textbf{First-order model properties correlate with planning performance, while zeroth-order error can be misleading.} Representative one-step predictions and derivatives along a multi-behavior trajectory. Standard MLPs can achieve slightly lower MAE but produce sharper, stiffer derivatives, whereas SNS-MLPs maintain smoother derivatives and yield
better closed-loop MPC performance.}
\label{fig: traj compare}
\end{figure*}

\noindent \textbf{Error assumptions.} We have shown that the dynamics residuals follow a Cauchy distribution more closely than a Gaussian distribution (Fig.~\ref{fig: empirical distributions}). Now, we investigate how the choice of likelihood affects training and control. In the Gaussian case, the Cauchy terms in~\eqref{eq: dynamics loss} and~\eqref{eq: estimator loss} are replaced with the mean Mahalanobis error (MME)
\begin{equation}
    \mathcal{L}_{\text{MME}}
    = \tfrac{1}{nN_\mathcal{D}}
      \sum_{x\in\mathcal{D}}
      \tfrac12 (x-\mu)^\top \Sigma^{-1}(x-\mu).
\end{equation}

The median and MAD are replaced by the mean and standard deviation for the normalization layers and loss calculations. We preserve the scaling for fairness instead of using the more common MSE (i.e., $\mathcal{L}_{\text{MME}}(x,\mu,I)$). Models trained with the heavy-tailed Cauchy likelihood achieve lower prediction error, faster and more stable convergence, and substantially lower control cost than those trained with Gaussian MLE  (Fig.~\ref{fig: mse vs cauchy}). This suggests that heavy-tailed residual models are better suited to train supervised models for legged robots, in our setting. Both setups use the same $c_{\text{ub}}$ and Lipschitz penalty weights $\lambda_3$ and $\lambda_4$. Yet, the dynamics models trained with Gaussian likelihoods struggle to maintain similar smoothness and error levels as the Cauchy-trained models.

\noindent\textbf{Exponential vs.\ softplus parameterization.} Fig.~\ref{fig: lip vs sns mpc} compares SNS-MLPs with Lipschitz MLPs~\cite{liu2022learning} to separate the effect of the parameterization from that of Lipschitz regularization itself. Both models share the architecture, objective (Eq.~\eqref{eq: dynamics loss}, $c_{\text{ub}}=10^4$), optimizer, learning rate, data collection, state estimator, and MPC; only the parameterization of $c_\ell$ differs. We test networks with $\approx$6M and $\approx$24M parameters (Appendix~\ref{ap: quadruped experiments}) and report the median cumulative cost over checkpoints between 25,000 and 100,000 training steps. For the smaller networks, SNS-MLPs reach the sensitivity budget about 2.4$\times$ sooner (20,400 vs.\ 49,600 steps), with lower rollout MAE at 100,000 steps (0.83 vs.\ 0.89) and lower cost (11.6 vs.\ 14.7).

For the larger networks, the Lipschitz MLP never reaches its budget and remains several orders of magnitude above it for most of training. Although this effectively unconstrained model attains the lowest rollout MAE of all models before it collapses ($\approx$0.75), it incurs the highest cost (89.5 vs.\ 11.8 for SNS-MLPs). Late in training, the softplus parameterization drives its input-layer constant to $\approx$0.01 and $c_{\max}/c_{\min}$ from 2 to $\approx 4\cdot10^3$ (Fig.~\ref{fig: lip vs sns mpc}d). Its rollout MAE then rises to 1.39, and only 14\% of trials succeed at the final checkpoint (i.e., avoid base or hip contact). The larger SNS-MLP reaches the budget within 28,200 steps and performs like the smaller one (MAE 0.87). These results suggest that Lipschitz regularization is what benefits MPC, while the parameterization determines whether that regularization is achieved as networks grow.



\subsection{Zero-Shot Generalization}\label{sec: zero-shot}
\noindent We now examine zero-shot generalization in more challenging control scenarios, with the goal of understanding when and why smooth neural surrogates succeed while standard neural networks fail. In particular, we aim to disentangle the effects of model smoothness, likelihood choice, solver type, and state estimation. To this end, each trained model type is evaluated after 100,000 training steps using both sampling-based and gradient-based MPC, with and without learned state estimation. We additionally compare smooth neural surrogates against early-stopped standard MLPs to assess whether observed performance differences arise primarily from reduced overfitting, from smoothness, or from their interaction. Early-stopped standard MLPs are frozen after 12,500 training steps, prior to the onset of any mean control performance degradation over 5~s evaluation episodes (Fig.~\ref{fig: smooth vs others}). We use SNS-MLPs with $c_\text{ub}=10^4$ and, as previously mentioned, DIAL-MPC~\cite{xue2025full} is used as a representative sampling-based method.

\begin{table}[t]
\caption{Dynamics Test Error for Different Models and Likelihoods}
\label{tab:MAE}

\noindent
\begin{minipage}[t]{0.3\columnwidth}
\centering
\vspace{-5pt}
\begin{tabular}{@{\hspace{2pt}}l@{\hspace{5pt}}l@{\hspace{2pt}}}
\hline
\textbf{Method} & \textbf{One-step MAE}\\
\hline
\noalign{\vskip 1pt}
MLP--Gaussian & $0.540\pm0.001$ \\
MLP--Cauchy (ES) & $0.545\pm0.001$ \\
\textbf{MLP--Cauchy}   & $\mathbf{0.476\pm0.001}$ \\
SNS--Gaussian & $0.604\pm0.001$ \\
SNS--Cauchy  & $0.571\pm0.001$ \\
\hline
\end{tabular}
\end{minipage}
\hfill
\begin{minipage}[t]{0.48\columnwidth}
\vspace{-12pt}
\footnotesize
\textbf{Notes.}
Cauchy likelihoods reduce error (\textbf{mean $\pm$ 95\% CI}) across architectures.
Smooth neural surrogates introduce bias under nonsmooth dynamics, but yield better performance in both sampling-based and gradient-based MPC. ES denotes early stopping at 12,500 training steps.
\end{minipage}

\end{table}

\begin{table*}[t]
\setlength{\tabcolsep}{5pt}
\caption{Model, Likelihood, and Optimizer Performance Across Sample Behaviors at Test-Time\label{tab:performance_methods}}
\vspace{-5pt}
\centering
\begin{tabular}{l@{\hspace{2pt}}c@{\hspace{6pt}}c@{\hspace{6pt}}c@{\hspace{6pt}}c@{\hspace{6pt}}c}
\hline
\noalign{\vskip 1pt}
\textbf{Method} &
\textbf{In-place trot } &
\textbf{Forward trot} ($1$ [m/s]) &
\textbf{Left trot} ($1$ [m/s]) &
\textbf{Forward gallop} ($1.5$ [m/s]) \\[0.5pt]
\hline
\noalign{\vskip 2pt}

MLP--Gaussian--DIAL
& $7.05\!\pm\!0.40$ / $7.16\!\pm\!0.41$ & $10.44\!\pm\!0.61$ / $10.78\!\pm\!1.12$ &  $44.80\!\pm\!15.72$ / $208.46\!\pm\!277.71$ & $24.54\!\pm\!0.95$ / $26.67\!\pm\!1.26$  \\

& $(5/5)$ / $(5/5)$
& $(5/5)$ / $(5/5)$
& $(0/5)$ / $(0/5)$
& $(5/5)$ / $(5/5)$
 \\[1pt]

MLP--Gaussian--GGN
& $9.62\!\pm\!2.40$ / $8.57\!\pm\!1.89$
& $5.15\!\pm\!3.19$ / $14.50\!\pm\!23.18$ & $301.00\!\pm\!74.40$ / $392.49\!\pm\!311.24$ & $33.35\!\pm\!3.62$ / $34.84\!\pm\!4.45$  \\

& $(5/5)$ / $(5/5)$
& $(5/5)$ / $(5/5)$
& $(0/5)$ / $(0/5)$
& $(0/5)$ / $(0/5)$
\\[1pt]

SNS--Gaussian--DIAL
& $6.21\!\pm\!0.19$ / $21.70\!\pm\!17.41$ & $8.50\!\pm\!0.54$ / $24.70\!\pm\!9.32$ & $203.78\!\pm\!313.91$ / $26.75\!\pm\!2.77$ & $30.53\!\pm\!4.27$ / $49.83\!\pm\!8.46$ \\
& $(5/5)$ / $(5/5)$
& $(5/5)$ / $(5/5)$
& $(4/5)$ / $(5/5)$
& $(3/5)$ / $(5/5)$
 \\[1pt]
 
SNS--Gaussian--GGN
& $2.09\!\pm\!0.01$ / $10.85\!\pm\!0.65$
& $3.87\!\pm\!0.03$ / $13.49\!\pm\!0.28$ 
& $4.67\!\pm\!0.06$ / $22.53\!\pm\!1.04$ 
& $17.13\!\pm\!2.10$ / $35.96\!\pm\!1.94$ \\

& $(5/5)$ / $(5/5)$
& $(5/5)$ / $(5/5)$
& $(5/5)$ / $(5/5)$
& $(5/5)$ / $(5/5)$
 \\[1pt]

MLP--Cauchy--DIAL
& $6.77\!\pm\!0.30$ / $11.34\!\pm\!1.84$
& $11.97\!\pm\!0.86$  / $15.35\!\pm\!3.23$ 
& $33.33\!\pm\!13.47$  / $27.67\!\pm\!9.60$ 
& $27.39\!\pm\!1.01$ / $33.91\!\pm\!6.86$
 \\
\quad  (early stopped)
& $(5/5)$ / $(5/5)$
& $(5/5)$ / $(5/5)$
& $(2/5)$ / $(4/5)$
& $(4/5)$ / $(3/5)$
 \\[1pt]

MLP--Cauchy--GGN
& $2.19\!\pm\!0.01$ / $3.19\!\pm\!0.05$ 
& $4.40\!\pm\!0.06$ / $5.59\!\pm\!0.16$
& $4.34\!\pm\!0.04$  / $5.37\!\pm\!0.02$ 
&  $23.02\!\pm\!2.81$ / $21.38\!\pm\!1.82$

 \\
\quad (early stopped)
& $(5/5)$ / $(5/5)$
& $(5/5)$ / $(5/5)$
& ${}^*$$(5/5)$ / $(5/5)$
& $(0/5)$ / $(2/5)$
 \\[1pt]
 
MLP--Cauchy--DIAL
& $6.29\!\pm\!0.26$
 / $6.15\!\pm\!0.21$
& $20.10\!\pm\!12.01$ / $31.68\!\pm\!56.19$ & $207.25\!\pm\!171.04$ / $174.35\!\pm\!290.23$ 
& $27.34\!\pm\!8.39$ / $54.49\!\pm\!46.58$ \\

& $(5/5)$ / $(5/5)$
& $(4/5)$ / $(1/5)$
& $(0/5)$ / $(0/5)$
& $(4/5)$ / $(2/5)$
 \\[1pt]

MLP--Cauchy--GGN
& $2.54\!\pm\!0.03$ / $2.57\!\pm\!0.03$
& $3.86\!\pm\!0.02$ / $3.90\!\pm\!0.03$
& $256.43\!\pm\!433.35$ / $158.78\!\pm\!76.79$
&  $53.27\!\pm\!14.94$ / $53.34\!\pm\!22.85$
 \\

& $(5/5)$ / $(5/5)$
& $(5/5)$ / $(5/5)$
& $(0/5)$ / $(0/5)$
& $(0/5)$ / $(0/5)$
 \\[1pt]

SNS--Cauchy--DIAL
& $5.93\!\pm\!0.24$ / $5.91\!\pm\!0.23$ & $7.62\!\pm\!0.29$ / $7.98\!\pm\!0.40$ & $15.66\!\pm\!1.40$ / $17.11\!\pm\!0.73$ & $24.94\!\pm\!1.07$ / $25.44\!\pm\!1.18$  \\

& $(5/5)$ / $(5/5)$
& $(5/5)$ / $(5/5)$
& $(5/5)$ / $(5/5)$
& $(5/5)$ / $(5/5)$
 \\[1pt]

\textbf{SNS--Cauchy--GGN}
& $\mathbf{1.87\!\pm\!0.01}$ / $\mathbf{1.99\!\pm\!0.02}$
& $\mathbf{3.46\!\pm\!0.03}$ / $\mathbf{3.69\!\pm\!0.03}$ & $\mathbf{3.99\!\pm\!0.02}$ / $\mathbf{4.10\!\pm\!0.01}$ & $\mathbf{15.35\!\pm\!0.11}$ / $\mathbf{15.65\!\pm\!0.18}$  \\
& $(5/5)$ / $(5/5)$
& $(5/5)$ / $(5/5)$
& $(5/5)$ / $(5/5)$
& $(5/5)$ / $(5/5)$
 \\[1pt]

\hline
\end{tabular}

\begin{minipage}{0.97\textwidth}
\vspace{2pt}
\textbf{Notes.} Each entry reports the cumulative cost and success rate over 11~second episodes (Top: \textbf{mean $\pm$ 95\% CI}. Bottom: \textbf{(success rate)}), shown as \textbf{GT / Est} corresponding to evaluation with ground-truth states and with learned state estimation, respectively. A trial is counted as \textbf{successful} if the robot avoids any undesired contact between the ground and the \textbf{base or hips} for the full episode. Method names follow the convention \textbf{NN--Likelihood--Optimizer}, where MLP denotes a standard MLP and SNS denotes our proposed smooth neural surrogate (SNS-MLP); Gaussian or Cauchy indicates the training likelihood; and DIAL or GGN denotes sampling-based DIAL-MPC \cite{xue2025full} and our gradient-based generalized Gauss--Newton (GGN) solver. Early stopping occurred at 12,500 training steps. During these trials, DIAL-MPC had a solve time of $\approx$11.5~ms per control step, while GGN-MPC required $\approx$8.0~ms per solve. \noindent \textbf{${}^*$ Right-trot note.}
For brevity, only left-trot results are reported for all methods. In right-trot,
MLP--Cauchy--GGN (early stopped) achieved success rates of $(3/5)$ / $(0/5)$, while
SNS--Cauchy--GGN achieved $(5/5)$ / $(5/5)$.
\end{minipage}
\end{table*}

\noindent\textbf{Model, likelihood, and solver performance.}
We begin by comparing test-time dynamics error across model architectures and likelihoods (Table~\ref{tab:MAE}).
Standard MLPs achieve the lowest one-step prediction error, with Cauchy likelihoods consistently improving performance relative to Gaussian losses. Smooth neural surrogates, by contrast, exhibit a modest increase in test error, reflecting the bias introduced by enforcing global smoothness on inherently nonsmooth contact dynamics.

Because the test dataset is generated using random actions and the training conditions are otherwise identical, one might expect a lower test error to correspond with improved control performance, particularly for sampling-based MPC. As we show next, this intuition does not hold. Among models with comparable prediction error, continuity and smoothness are strong predictors of reliable and generalizable control.

Table~\ref{tab:performance_methods} reports cumulative cost and success rates across representative locomotion tasks for all combinations of model, likelihood, solver, and state estimation. Several trends emerge. Standard MLPs trained to completion can execute some relatively stable behaviors, such as in-place or forward trotting, but typically incur higher cost and fail outright on more challenging tasks, including lateral trotting. In multiple cases, standard MLPs also perform substantially worse under gradient-based optimization than under sampling-based MPC, highlighting their sensitivity to poorly conditioned or noisy derivatives. Early-stopped standard MLPs improve performance on some of the more challenging behaviors relative to fully trained models, but this comes at the expense of increased prediction error, leading to degraded performance on some simpler tasks. Moreover, early stopping alone does not prevent catastrophic failures, as early-stopped models still fail at right-trotting.

In contrast, smooth neural surrogates reliably improve performance across both solver classes. When paired with our generalized Gauss--Newton solver, SNS-MLPs achieve uniformly low costs and perfect success rates across all evaluated behaviors. Cauchy likelihoods further enhance robustness, which is particularly noticeable when using state estimation. Moreover, performance using learned state estimates closely matches that obtained with ground-truth states when using SNS-based dynamics and estimation.

Taken together, these results demonstrate that despite exhibiting higher test error, smooth neural surrogates dramatically outperform standard MLPs in both sampling-based and gradient-based control. This discrepancy indicates that model error fails to capture properties critical for closed-loop planning, which is consistent with prior work~\cite{lutter2021learning}. The observed behavior suggests a form of overfitting (for control purposes) tied to local nonsmoothness and derivative instability, phenomena that are not detectable through zeroth-order test error alone and are not fully mitigated by early stopping. Accordingly, we observe no setting in which performance degrades when SNS-MLPs and Cauchy likelihoods are combined. At worst, performance consistently improves by about 10–50\%, and with more complex control tasks, these models enable reliable execution (0/5$\rightarrow$5/5 success), yielding $\approx$2–50$\times$ reductions in cost.

\noindent\textbf{Overfitting, stiff gradients, or both?}
In Section~\ref{sec: smooth neural networks}, we showed in a 2-D shape interpolation example that local nonsmoothness often emerges in regions of sparse data, and it can be linked, in a practical sense, to overfitting. The failure of standard MLP dynamics in several locomotion tasks raises an analogous question here: \emph{do smooth neural surrogates succeed because they attenuate stiff contact dynamics, or because they suppress overfitting-induced local nonsmoothness?}

The test errors reported in Table~\ref{tab:MAE} do not suggest that standard MLPs should perform worse in control; if anything, their lower MAE would predict the opposite. To better understand this discrepancy, we analyze model predictions along a challenging trajectory generated using SNS-MLP dynamics and GGN-MPC (Fig.~\ref{fig: traj compare}). This trajectory contains multiple gaits and transitions that standard MLPs consistently fail to reproduce when deployed within an MPC loop.

Along this trajectory, the fully trained standard MLP still attains a lower MAE than the SNS-MLP and visually captures more sharp transients and contact events. At the level of zeroth-order predictions, both models appear broadly similar, except for these rapid changes. The key distinction instead emerges at first order: the derivatives of the SNS-MLP remain smooth, bounded, and well-conditioned, whereas those of the fully trained MLP are noisy, stiff, and highly variable. The early-stopped MLP exhibits noticeably smoother derivatives than the fully trained model, consistent with the intuition that continued training leads to sharper dynamics. However, in the absence of an explicit smoothness constraint, early stopping alone does not prevent the emergence of some locally nonsmooth or noisy gradients, and this comes at the cost of substantially higher prediction error.

\begin{figure}
    \centering
    \includegraphics[width=\linewidth]{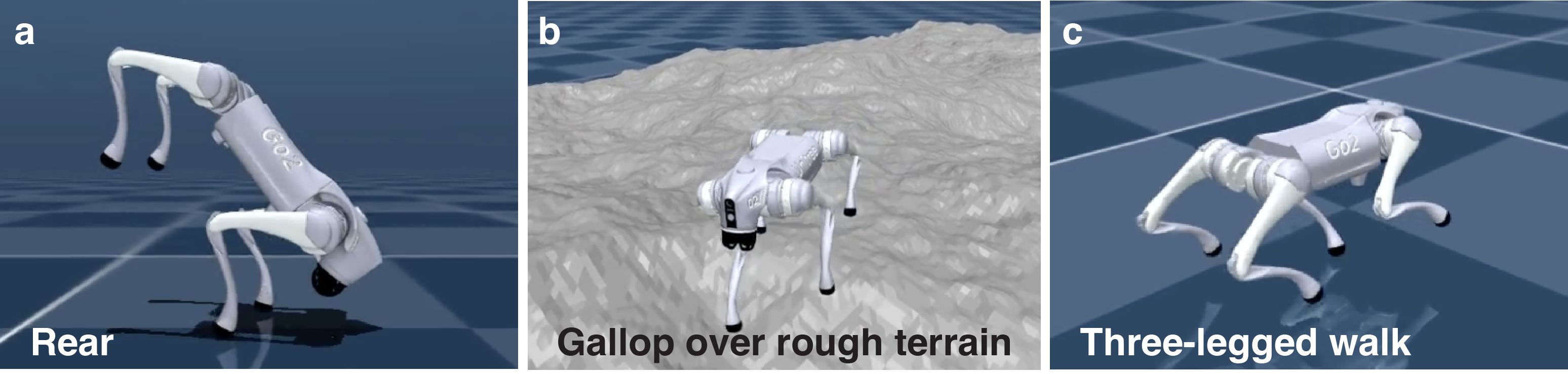}
    \caption{\textbf{High-coordination (blind) behaviors synthesized without task or environment specific training using smooth neural surrogates and GGN-MPC. a}, Stable balance in the rearing position. \textbf{b}, Rotary gallop over rough terrain at 1.5 meters per second. \textbf{c}, Three-legged locomotion.}
    \label{fig: sim only}
\end{figure}

These observations suggest that smooth neural surrogates provide two complementary advantages for control.
First, they mitigate stiffness arising from contact dynamics, yielding gradients that are better suited for optimization.
Second, by enforcing global regularity, they suppress local nonsmoothness in regions of the state–action space that are subject to overfitting or rife with local optima. The combination of these effects enables accurate tracking, coordinated whole-body control, and robustness against catastrophic failures, ultimately supporting reliable zero-shot generalization to new behaviors and terrains (Fig.~\ref{fig: sim only}).


\subsection{Unlocking the Benefits of Gradient-Based MPC}\label{sec: unlocking gradient-based}
\noindent In the previous subsection, we showed that smooth neural surrogates stabilize sampling- and gradient-based MPC. In this section, we examine the practical advantages of gradient-based MPC enabled by our smooth neural dynamics. We compare GGN-MPC to DIAL-MPC \cite{xue2025full} under matched MPC formulations, spline-parameterized controls, and identical costs, dynamics, and state estimators, evaluating both solve time and closed-loop performance over full episodes.

\begin{figure}
    \centering
    \includegraphics[width=\linewidth]{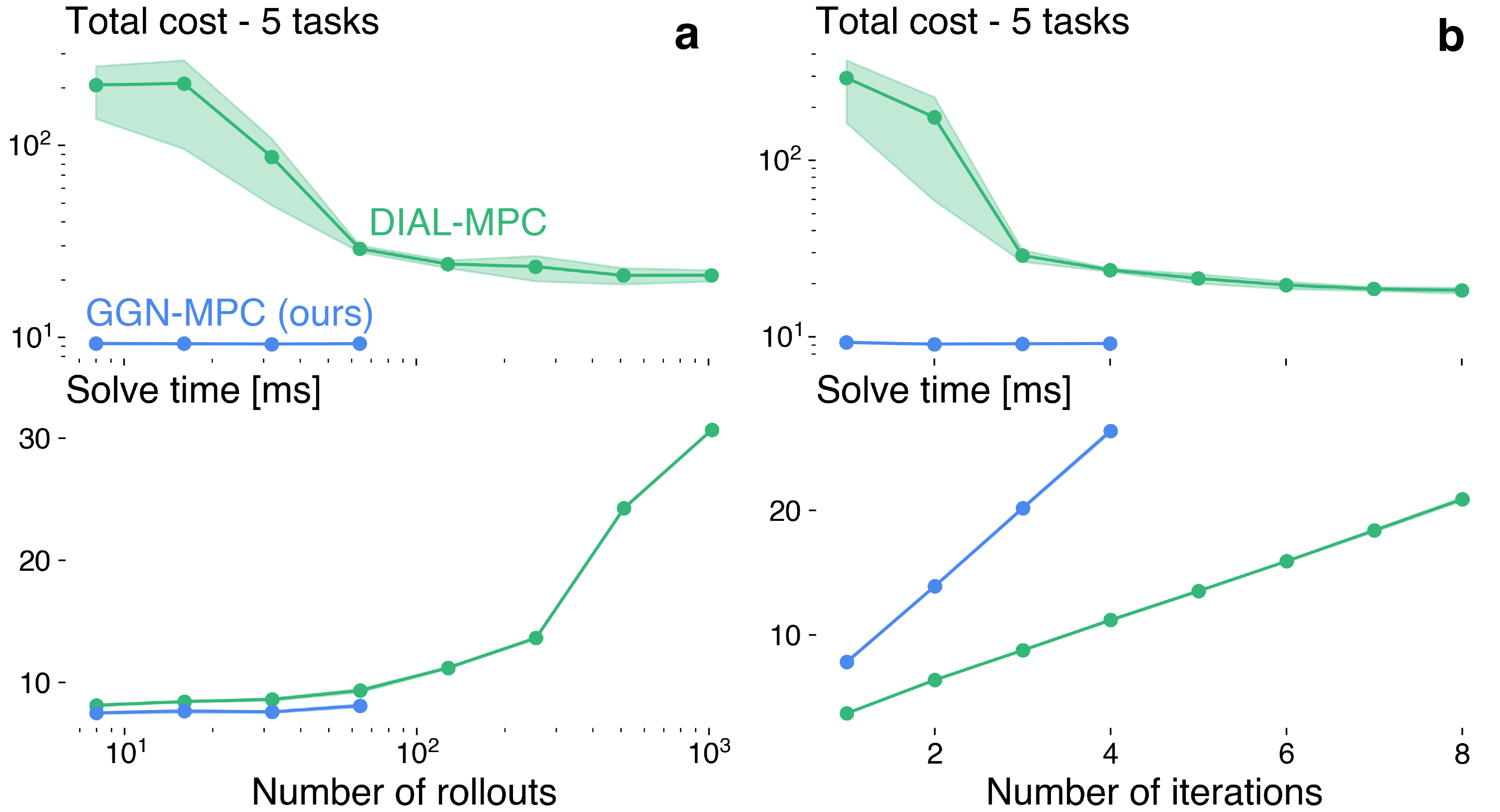}
    \caption{\textbf{Convergence and computational efficiency: gradient-based vs. sampling-based MPC with SNS.} Overall we observe \textbf{$\approx$2$\times$ lower cost at equal solve times} when using a fixed, modest, number of action knots ($k=5$). \textbf{a}, With a fixed iteration count (4 for DIAL-MPC, 1 for GGN-MPC), DIAL-MPC performance depends strongly on rollout count, while GGN-MPC remains consistently low-cost. \textbf{b}, With a fixed rollout count (128 vs. 16), GGN-MPC shows diminishing returns from additional iterations, whereas DIAL-MPC requires more iterations and still fails to reach comparable cost. }
    \label{fig: ggn 0}
\end{figure}

\noindent\textbf{Overall performance and sensitivity to hyperparameters.} GGN-MPC and DIAL-MPC share the number of rollouts as a key forward-pass hyperparameter. DIAL-MPC samples candidates from an annealed isotropic Gaussian, while GGN-MPC evaluates deterministic proposals along the descent direction from the backward pass. Although sampling-based MPC improves with more rollouts, Fig.~\ref{fig: ggn 0} shows that GGN-MPC selects far more informative candidates, achieving low cost with an order of magnitude fewer evaluations. As a result, its cumulative control cost remains significantly lower across all tested numbers of rollouts. GGN-MPC also gains little from additional iterations, whereas DIAL-MPC depends heavily on repeated solves. A single DIAL-MPC iteration is faster, but multiple iterations are required for convergence, reducing overall efficiency. Overall, we observe about 2$\times$ lower cumulative cost at equal, and realistic, solve times.

\begin{figure}
    \centering
    \includegraphics[width=\linewidth]{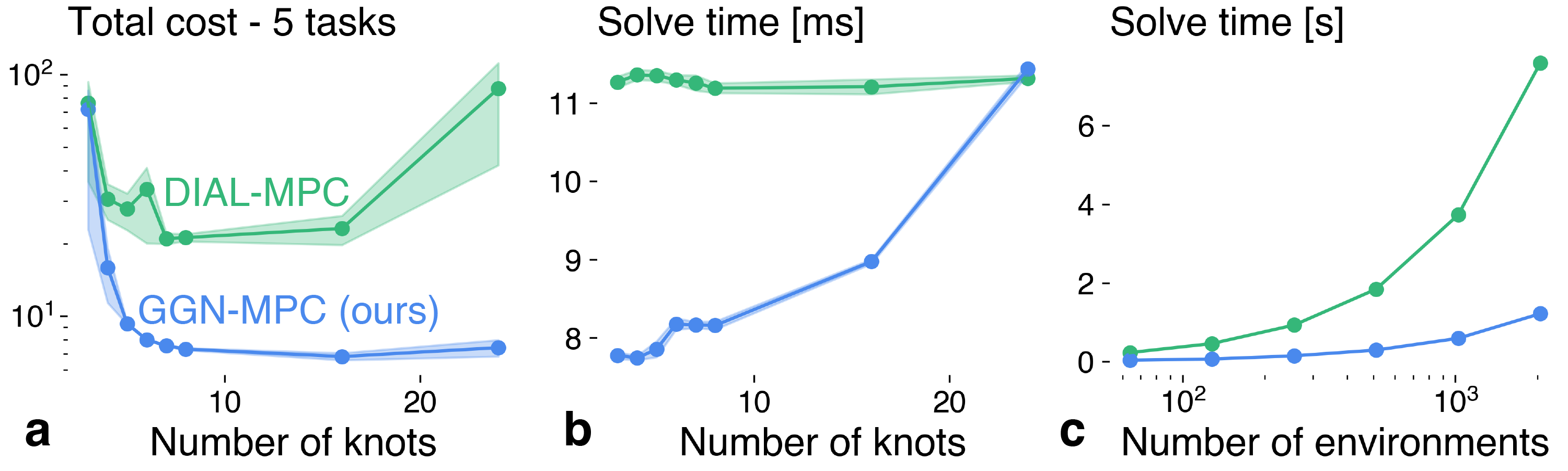}
    \caption{\textbf{Scalability: gradient-based vs. sampling-based MPC with SNS.} 
    \textbf{a}, GGN-MPC maintains low cost across action-trajectory resolutions; DIAL-MPC fails as dimensionality increases (\textbf{$\approx$10$\times$ lower cost} at 24 knots).
    \textbf{b}, GGN-MPC solve time grows quadratically with knots yet remains efficient and achieves much lower mean cost.
    \textbf{c}, In parallel environments, sampling-based MPC scales poorly due to large rollout requirements, while GGN-MPC remains more efficient (\textbf{$\approx$7$\times$ faster} at 1024 environments).}
    \label{fig: ggn 2}
\end{figure}

\noindent\textbf{Scalability.} 
We compare scalability along two axes: the dimensionality of the decision vector and parallelization across environments during RL training. In our setup, DIAL-MPC is run for four iterations with 128 rollouts per iteration, whereas GGN-MPC uses a single iteration with only 16 rollouts. As shown in Fig.~\ref{fig: ggn 2}, both methods initially benefit from increasing the action-trajectory resolution, reaching peak performance at roughly ten spline knots ($\approx$120 decision variables). Beyond this point, DIAL-MPC degrades rapidly. This is an instance of the curse of dimensionality that is characteristic of sampling-based methods. GGN-MPC exhibits quadratic growth in solve time due to Cholesky factorization but remains faster at all but the highest resolutions. GGN-MPC also sustains very low costs with high-dimensional decision variables where DIAL-MPC fails, despite using only a single iteration. Overall, this corresponds to $\approx$10$\times$ lower cumulative cost at about 300 decision variables.

Parallelization is the cornerstone of large-scale RL. Our JAX implementation parallelizes the optimizers across environments using vectorized mapping $\texttt{vmap}(\texttt{MPC}_\theta)$. Because GGN-MPC requires significantly fewer rollouts and thus less computation per agent, it scales more effectively in GPU-based parallel training, as illustrated in Fig.~\ref{fig: ggn 2}. This results in about 7$\times$ faster solves at 1024 environments.

\begin{figure}
    \centering
    \includegraphics[width=\linewidth]{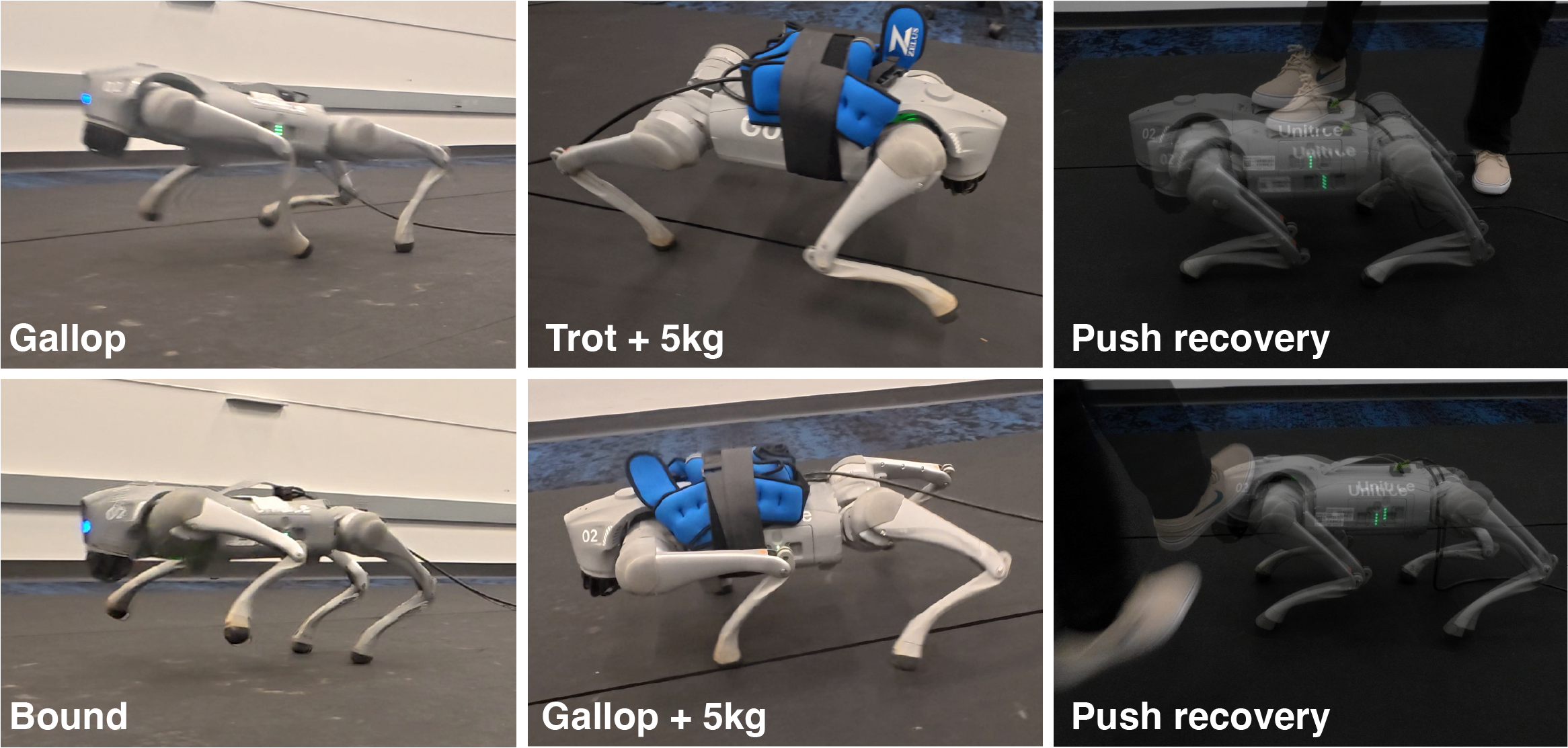}
    \caption{\textbf{Indoor hardware experiments.} Bounding, galloping, load carrying, and push recovery with the same frozen models.}
    \label{fig: bounding type}
\end{figure}

\begin{figure}
    \centering
    \includegraphics[width=\linewidth]{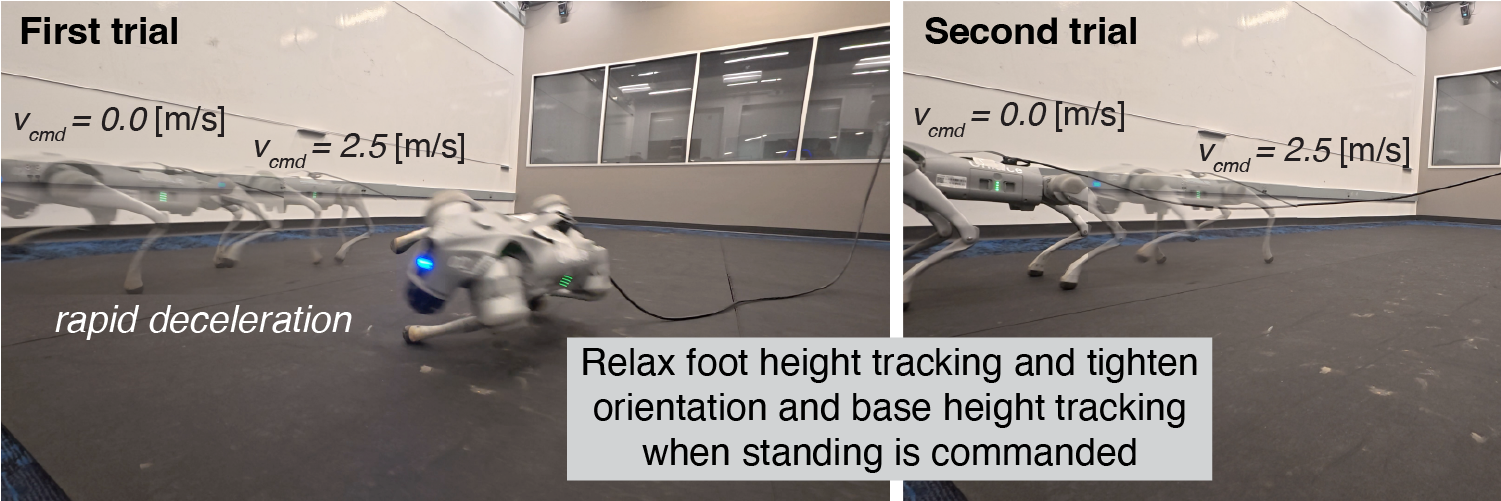}
    \caption{\textbf{Cost tuning on hardware.} The standing controller, which is engaged when no velocity is commanded, is not tested in simulation; instead, we begin with the trotting cost using a zero commanded foot height. This formulation fails to decelerate the robot at high speeds because it over-penalizes stepping and under-penalizes orientation and height tracking, causing the robot to fall by not taking an extra stabilizing step. After adjusting these weights, the controller achieves stable deceleration from sprinting.}
    \label{fig: online tuning}
\end{figure}
\begin{figure*}
    \centering
    \includegraphics[width=\linewidth]{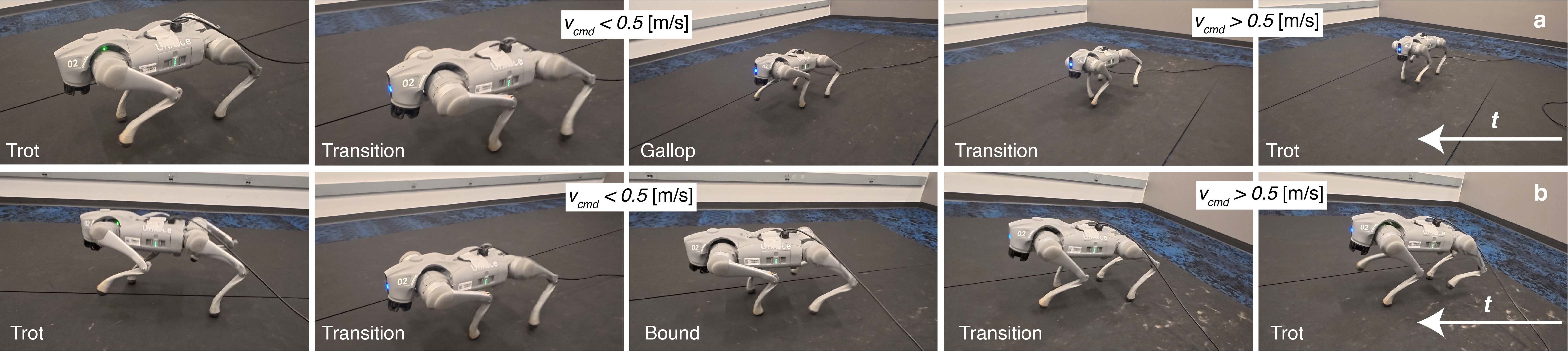}
    \caption{\textbf{Swapping costs and constraints in real time.} At each control step, the MPC solves a distinct objective. Low velocity commands use the default trotting cost, while crossing the 0.5\,m/s threshold triggers the transition cost, which provides no explicit gait reference and allows the controller to plan its own steps between gaits. \textbf{a}, For velocity commands above 0.5\,m/s, the controller blends the transition cost with the galloping objective using an exponential moving average. \textbf{b}, Similarly, velocity commands above 0.5\,m/s blend the transition cost with the bounding objective.
}
    \label{fig: gait transitions}
\end{figure*}
 
\subsection{Hardware Deployment}
\noindent Using a single dynamics model and state estimator trained offline and frozen at test time, we generate a wide range of whole-body behaviors by modifying the costs, constraints, and commands online (Fig.~\ref{fig: teaser}). Our use of domain randomization during training, together with signed-distance state augmentation, enables robust terrain- and body-aware planning. As shown in Fig.~\ref{fig: whole-body}, our MPC maintains accurate collision awareness and contact reasoning across uneven slopes, slippery surfaces, deep debris, and reliably recovers from disturbances such as falls.

Indoors, the same dynamics and estimator produce diverse locomotion patterns—including bounding, galloping, load-carrying, and multi-directional push recovery (Fig.~\ref{fig: bounding type}). We take advantage of the task-level flexibility of MPC, refining behavior quality through cost tuning to correct for failure modes such as high-speed deceleration (Fig.~\ref{fig: online tuning}). Because we explicitly incorporate behavior stitching (Appendix~\ref{ap: quadruped experiments}), the controller transitions smoothly between gaits with different costs and constraints (Fig.~\ref{fig: gait transitions}), without explicitly training for this behavior. Together, these results demonstrate that our offline-learned smooth neural dynamics, when combined with online MPC, enable flexible, robust, and versatile behavior generation in the real world.

\begin{table*}[t]
\setlength{\tabcolsep}{4pt}
\caption{Cost Terms and Weights Used for Neural MPC on the Go2 Quadruped Across Behaviors\label{tab:costs}}
\vspace{-5pt}
\centering
\begin{tabular}{cccccccccc}
\hline
\noalign{\vskip 1pt}
\multicolumn{2}{c}{\textbf{Cost Term}} &
\textbf{Training} &
\textbf{Trot} &
\textbf{Bound/Gallop} &
\textbf{Pace} &
\textbf{Tripod} &
\textbf{Transition} &
\textbf{Stand} \\[0.25pt]
\hline
\noalign{\vskip 1pt}
\textbf{Name} & \textbf{Equation} &
\multicolumn{6}{c}{\textbf{Weight}} \\[-0.25pt]
\hline
\noalign{\vskip 2pt}

Base orientation tracking
& $\|\log(\hat R_\mathcal{W}^{\!\top} R^{\text{cmd}}_\mathcal{W})^\vee\|^2$
& $1$ & $1$
& $0.65$
& $1$ & $0.7$ & EMA & $2$ \\[1pt]

Base height tracking
& $\|\hat z - z^{\text{cmd}}\|^2$
& $5$ & $5$
& $1.25$
& $5$ &$0.5$ & EMA & $6$ \\[1pt]

Base linear velocity tracking
& $\|\hat v_\mathcal{B} - v^{\text{cmd}}_\mathcal{B}\|^2$
& $3\!\cdot\!10^{-2}$ & $5\!\cdot\!10^{-2}$
& $5\!\cdot\!10^{-2}$
& $5\!\cdot\!10^{-2}$ & $5\!\cdot\!10^{-2\dagger\dagger}$  & EMA & $5\!\cdot\!10^{-2}$ \\[1pt]

Base angular velocity tracking
& $\|\hat\omega_\mathcal{B} - \omega^{\text{cmd}}_\mathcal{B}\|^2$
& $10^{-3}$ & $10^{-3}$
& $5\!\cdot\!10^{-4}$
& $10^{-3}$ & $5\!\cdot\!10^{-4\ddagger}$ & EMA & $10^{-3}$ \\[1pt]

Gait / foot height tracking
& $\|\hat\phi_{\text{foot}} - \phi^{\text{cmd}}_{\text{foot}}\|^2$
& $0$ & $2$
& $7$
& $4$ & $7^\dagger$ & Prev. & $0.5$ \\[1pt]

Nominal joint position penalty
& $\|\hat q_J - q^{\text{nom}}_J\|^2$
& $10^{-2}$ & $10^{-2}$
& $10^{-2}$
& $10^{-2}$ & $10^{-2}$ & $5\!\cdot\!10^{-4}$ & $10^{-2}$\\[1pt]

Joint velocity penalty
& $\|\hat v_J\|^2$
& $10^{-8}$ & $10^{-2}$
& $5\!\cdot\!10^{-8}$
& $10^{-2}$ & $5\!\cdot\!10^{-8}$ & $10^{-6}$ & $10^{-2}$ \\[1pt]

Torque penalty
& $\|\hat{\tau}\|^2$
& $2\!\cdot\!10^{-6}$ & $4\!\cdot\!10^{-6}$
& $4\!\cdot\!10^{-6}$
& $4\!\cdot\!10^{-6}$ & $4\!\cdot\!10^{-6}$ & $4\!\cdot\!10^{-6}$ & $4\!\cdot\!10^{-6}$ \\[1pt]

Pos. mech. work penalty
& $\|\max(0,\hat{\tau}\cdot\hat v_J)\|^2$
& 0 & $10^{-6}$
& $10^{-6}$
& $10^{-6}$ & $10^{-6}$ & EMA & $10^{-6}$ \\[1pt]

Drift penalty$^{*}$
& $\|\!\sum_{t=0}^{T}(\hat v_{xy}(t)-v^{\text{cmd}}_{xy}(t))dt\|^2$
& $0$ & $10^{-5}$
& $10^{-5}$
& $10^{-5}$ & $10^{-5}$ & EMA & $10^{-5}$\\[2pt]

\hline
\end{tabular}
\begin{minipage}{0.97\textwidth}
\vspace{2pt}
\textbf{Notes.} All costs are stagewise unless marked with $^{*}$, which denotes a trajectory-level (global) cost. The global cost is used during the MPC solve but excluded when evaluating the closed-loop control cost. $^\dagger$: FR foot has a weight of $35$. $^{\dagger\dagger}$: Velocity $v_y$ has a weight of $25$. $^{\ddagger}$: Angular velocity $\omega_z$ has a weight of $25$. EMA indicates that an exponential moving average of the previous task weights and the next commanded task weights are used to determine the current value.
\end{minipage}
\end{table*}

\section{Conclusion}
\noindent This work examined whether learned dynamics models can support reliable, task-flexible MPC for legged robots despite the discontinuities and stiffness induced by contact. We showed that standard neural networks reproduce many of the same numerical pathologies that hinder classical trajectory optimization under contact, while introducing additional failure modes due to local nonsmoothness and overfitting. These observations motivated the smooth neural surrogate MLP, a lightweight yet scalable architecture with tunable Lipschitz bounds, paired with a heavy-tailed training objective that better reflects the impulse-like distribution of model errors observed in legged robots.

Across controlled simulation studies and hardware experiments, smooth neural surrogates yielded substantially more informative derivatives and improved regularity in the learned dynamics, benefiting both sampling-based and gradient-based optimization. When combined with a smooth learned estimator and domain randomization, a single frozen model was sufficient to synthesize a wide range of whole-body behaviors at test time, including locomotion over uneven terrain and multi-gait transitions, without behavior-specific training or finetuning.

Taken together, our results suggest that relatively simple architectural and training choices can substantially improve the compatibility between learned dynamics and optimization-based control. This provides a practical and scalable alternative to hand-designed models and standard model-based RL pipelines and points toward more flexible, general-purpose controllers for real-world legged robots.

\noindent \textbf{Limitations and future work.} While effective for quadruped locomotion, several extensions remain open. Applying smooth neural surrogates to contact-rich manipulation and humanoid robots may enable more general whole-body control. Incorporating smooth surrogates into policy learning pipelines could stabilize gradient-based updates by providing better-conditioned or adaptive model derivatives. Integrating raw perceptual inputs such as vision, depth, or tactile sensing into both the dynamics and estimator models would further improve terrain awareness and planning under partial observability. Although Lipschitz constraints have been proposed for more expressive architectures such as CNNs and transformers, their practical benefits in high-capacity networks for robotics and control remain largely unexplored. Moreover, higher-order smoothness constraints may prove useful in scenarios like differentiable optimization or Newton-based optimization, where second-order gradients are used. These directions suggest that smooth neural surrogates may serve as modular building blocks that enhance a broad class of scalable, contact-rich learning and control methods.

\section*{Acknowledgments}
\noindent This work was supported by the National Science Foundation Graduate Research Fellowship, by ARO under award W911NF2410405, and by DARPA TIAMAT program under award HR00112490419.
{\appendices

\section{Softplus versus Exponential Parameterization of Lipschitz Constants}\label{ap:softplus-exp-proof}

\noindent Under the softplus parameterization, $C_{\mathrm{sp}}(\theta_c)=\prod_{\ell}\operatorname{softplus}(\theta_c^{(\ell)})$, the derivative with respect to $\theta_c^{(j)}$ is
\begin{equation}
\label{eq:softplus_prod_grad}
\frac{\partial C_{\mathrm{sp}}}{\partial \theta_c^{(j)}}
=
\sigma(\theta_c^{(j)})
\prod_{\ell \neq j}
\operatorname{softplus}(\theta_c^{(\ell)})
=
\frac{ C_{\mathrm{sp}}(\theta_c)\,
\sigma(\theta_c^{(j)})
}{
\operatorname{softplus}(\theta_c^{(j)})
},
\end{equation}
where $\sigma(z)=(1+\exp(-z))^{-1}$ is the sigmoid function. Suppose the training objective contains a term $\mathcal{L}(C_{\mathrm{sp}})$ that penalizes large values of the product, with $\mathcal{L}'(C_{\mathrm{sp}})>0$. Then
\begin{equation}
\frac{\partial \mathcal{L}(C_{\mathrm{sp}})}
{\partial \theta_c^{(j)}}
=
\mathcal{L}'(C_{\mathrm{sp}})
C_{\mathrm{sp}}(\theta_c)
\frac{
\sigma(\theta_c^{(j)})
}{
\operatorname{softplus}(\theta_c^{(j)})
}.
\end{equation}
Thus, all layerwise gradients share the same global factor
$\mathcal{L}'(C_{\mathrm{sp}}) C_{\mathrm{sp}}(\theta_c)$, and their relative magnitudes are controlled by
\begin{equation}
r(\theta)
=
\frac{\sigma(\theta)}{\operatorname{softplus}(\theta)}
=
\frac{d}{d\theta}
\log \operatorname{softplus}(\theta).
\end{equation}

The same conclusion holds if the penalty is applied to $\log C$ rather than $C$: for softplus, $\partial \log C_{\mathrm{sp}}/\partial \theta_c^{(j)}=r(\theta_c^{(j)})$, whereas for the exponential parameterization this derivative is identically one.

\medskip
\noindent \textit{Proposition.} Let $r(\theta)=\sigma(\theta)/\operatorname{softplus}(\theta)$. Then $r$ is strictly decreasing in $\theta$. Consequently, for a gradient descent step on $\mathcal{L}(C_{\mathrm{sp}})$, if $\theta_c^{(a)} < \theta_c^{(b)}$, then the layer with the smaller trainable parameter $\theta_c$, equivalently the smaller $c_\ell$, receives a larger downward update in $\theta$-space: $|\Delta \theta_c^{(a)}|>|\Delta \theta_c^{(b)}|$.
Moreover, the separation between the two trainable parameters in $\theta$-space increases after the update.

\medskip
\noindent \textit{Proof.}
First, differentiate $r(\theta)$. Since $\operatorname{softplus}'(\theta)=\sigma(\theta)$, $\sigma'(\theta)=\sigma(\theta)(1-\sigma(\theta))$, we have
\begin{equation}
r'(\theta)
=
\frac{\sigma(\theta)}
{
\operatorname{softplus}(\theta)^2
}
\left[
(1-\sigma(\theta))
\operatorname{softplus}(\theta)
-
\sigma(\theta)
\right].
\end{equation}
Let $x=\exp(\theta)>0$. Then, $\sigma(\theta)=x/(1+x)$, $1-\sigma(\theta)=1/(1+x)$, and $\operatorname{softplus}(\theta)=\log(1+x)$.
Therefore,
\begin{equation}
(1-\sigma(\theta))
\operatorname{softplus}(\theta)
-
\sigma(\theta)
=
\frac{\log(1+x)-x}{1+x}
<0,
\end{equation}
because $\log(1+x)<x$ for all $x>0$. Hence $r'(\theta)<0$, so $r$ is strictly decreasing.

Now consider one gradient descent step with step size $\eta>0$:
\begin{equation}
\theta_{c,+}^{(j)}
=
\theta_c^{(j)}
-
\eta\,
\mathcal{L}'(C_{\mathrm{sp}})
C_{\mathrm{sp}}(\theta_c)
r(\theta_c^{(j)}).
\end{equation}
If $\theta_c^{(a)}<\theta_c^{(b)}$, then since $r$ is strictly decreasing, $r(\theta_c^{(a)})>r(\theta_c^{(b)})$. Thus,
\begin{equation}
\begin{aligned}
\underbrace{\eta\,
\mathcal{L}'(C_{\mathrm{sp}})
C_{\mathrm{sp}}(\theta_c)
r(\theta_c^{(a)})}_{|\Delta \theta_c^{(a)}|}
>
\underbrace{\eta\,
\mathcal{L}'(C_{\mathrm{sp}})
C_{\mathrm{sp}}(\theta_c)
r(\theta_c^{(b)})}_{|\Delta \theta_c^{(b)}|}
.
\end{aligned}
\end{equation}
Furthermore,
\begin{equation}
    \begin{aligned}
    \theta_{c,+}^{(b)}-\theta_{c,+}^{(a)}
&=
\theta_c^{(b)}-\theta_c^{(a)} \\
&+
\eta\,
\mathcal{L}'(C_{\mathrm{sp}})
C_{\mathrm{sp}}(\theta_c)
\left[
r(\theta_c^{(a)})-r(\theta_c^{(b)})
\right] \\
\end{aligned}
\end{equation}
So, $\theta_{c,+}^{(b)}-\theta_{c,+}^{(a)}>\theta_c^{(b)}-\theta_c^{(a)}$. This means the smaller parameter is pushed downward more aggressively, and the gap between small and large trainable parameters increases in $\theta$-space.
\hfill $\square$

\medskip
This shows that, when considering the product penalty in isolation, the softplus parameterization induces an asymmetric shrinkage effect in parameter space. A penalty on the product does not reduce all layerwise constants uniformly. Instead, the softplus parameterization preferentially suppresses the already-small parameters while leaving larger parameters comparatively less affected. Because the analysis does not model the task loss, it identifies a mechanism that can lead to collapse rather than proving collapse; Section~\ref{sec: convergence} tests it in full training.

For example, if one layer has $\theta=0$ and another has $\theta=10$, then $r(0)=\tfrac{1/2}{\log 2}\approx 0.72$, $r(10)=\sigma(10)/\operatorname{softplus}(10)\approx 0.10$.
Hence, under the same product penalty, the layer with $\theta=0$ receives a downward update more than seven times larger than the layer with $\theta=10$. Thus, under the product-penalty gradient alone, the softplus parameterization can amplify layerwise imbalance: layers with small constants are driven smaller more quickly, while layers with large constants are comparatively less sensitive to the regularizer.

To avoid this asymmetric shrinkage, we use the exponential parameterization $c_\ell=\exp(\theta_c^{(\ell)})$. In this case, $C_{\exp}(\theta_c)=\prod_{\ell}\exp(\theta_c^{(\ell)})=\exp(\sum_\ell \theta_c^{(\ell)})$ and $\partial C_{\exp}/\partial \theta_c^{(j)}=C_{\exp}(\theta_c)$. Therefore, $\partial \mathcal{L}(C_{\exp})/\partial \theta_c^{(j)}=\mathcal{L}'(C_{\exp})\,C_{\exp}(\theta_c)$, which is independent of $j$. Under gradient descent, $\theta_{c,+}^{(j)}=\theta_c^{(j)}-\eta\,\mathcal{L}'(C_{\exp})C_{\exp}(\theta_c)$, so every layerwise parameter is shifted downward by the same amount. Equivalently, $\log C_{\exp}(\theta_c)=\sum_\ell \theta_c^{(\ell)}$, $\partial \log C_{\exp}/\partial \theta_c^{(j)}=1$.
Thus, the exponential parameterization gives a balanced log-space geometry for the product constraint: each layer contributes equally to the logarithm of the global Lipschitz bound. Unlike the softplus parameterization, it does not preferentially shrink already-small layerwise constants, and therefore avoids amplifying imbalance across layers under the product-penalty gradient.
\section{Smooth Neural Surrogate Experiments}\label{ap: sns experiments}

\noindent All networks in this section use six layers with softplus activations, MSE loss and hidden layers with $3x_0$ units, where $x_0$ denotes the respective input dimension, and are trained with Adam ($\beta_1=0.9$, $\beta_2=0.999$). Each experiment is run with 3 random seeds, and final results are summarized in Table~\ref{tab:toy}.

\begin{table}[t]
\caption{Final Results of the Smooth Neural Surrogate Experiments}
\label{tab:toy}
\centering
\footnotesize
\setlength{\tabcolsep}{1.7pt}
\begin{tabular}{lcccc}
\hline
 & \multicolumn{2}{c}{\textbf{MSE}} & \multicolumn{2}{c}{\textbf{Smallest} $c_\ell$} \\
\textbf{Problem} & \textbf{Lip.\ MLP} & \textbf{SNS} & \textbf{Lip.\ MLP} & \textbf{SNS} \\
\hline
Shape ($\lambda\!=\!10^{-1}$) & $0.16{\pm}0.04$ & $(7.1{\pm}4.9)\!\cdot\!10^{-5}$ & $\leq$0.007 & 0.67--0.75 \\
Shape ($\lambda\!=\!10^{-3}$) & $0.014{\pm}0.009$ & $(1.2{\pm}1.0)\!\cdot\!10^{-5}$ & 0.009--0.029 & 0.42--0.56 \\
1-D piecewise & $0.183{\pm}0.003$ & $0.0055{\pm}0.0002$ & 0.003--0.005 & 0.07--0.08 \\
Particle--mass & $0.152{\pm}0.005$ & $0.039{\pm}0.006$ & 0.007--0.008 & 0.40 \\
\hline
\end{tabular}
\begin{minipage}{0.97\columnwidth}
\vspace{2pt}
\footnotesize
\textbf{Notes.} MSE is reported as mean $\pm$ s.d.\ over 3 seeds; the smallest layerwise constant at the end of training is reported as the range over seeds.
\end{minipage}
\end{table}

\noindent\textbf{2-D shape interpolation.}
 The standard MLP uses just an MSE loss, and the SNS and Lipschitz MLPs include a smoothness constraint with $c_{\text{ub}} = 8.0$ and $\lambda\in\{10^{-1},10^{-3}\}$ in addition to MSE. Our dataset contains 500K uniformly sampled points per shape. All networks are trained for 200 epochs with a learning rate of 0.003.

\noindent\textbf{1-D nonsmooth functions.}
For Fig.~\ref{fig:funky nonsmooth}, we train on 15K uniformly sampled $(x,y)$ pairs with $y \approx f_\theta(x)$ using an MSE loss for 1000 epochs with $c_{\text{ub}} = 1.35$, $\lambda = 0.1$, and a learning rate of 0.01. 

\noindent\textbf{Smoothing particle--mass contact dynamics.}
We generate 500 trajectories (6 s at 50 Hz) from random initial states $x_0 = [q_0, v_0] \sim [\mathcal{U}(0.1, 4.0), \mathcal{U}(-5.0, 5.0)]$ under sinusoidal forcing $u_t = 2g \sin(2\pi \omega t)$, $\omega \sim \mathcal{U}(0.1, 3.0)$. The dynamics follow $\ddot{q} = -g + u$ with inelastic collisions at $q=0$. Models are trained with a single-step MSE loss $x_{t+1} \approx x_t + f_\theta(x_t, u_t)$ for 500 epochs at a learning rate of $10^{-3}$, with $c_{\mathrm{ub}}=50$ and $\lambda=0.2$. 

\section{Quadruped Experiments}
\label{ap: quadruped experiments}

\begin{algorithm}[t]
\caption{Learning legged MPC and state estimation with smooth neural surrogates}
\label{alg:full-training}
\footnotesize
\begin{algorithmic}[1]
\REQUIRE Replay buffer $\mathcal{D}$; smoothness budgets; horizon $T$; history length $H$; episode length $T_{\text{episode}}$; domain randomization, command, and state reset distributions; actuator net $A_\psi$
\STATE Initialize SNS dynamics model $\hat{\mu}_\theta$, SNS estimator $\bar \mu_\zeta$, and simulator
\STATE Compute median and MAD over initial $\mathcal{D}$ to set dispersion $\hat{\Sigma}_\mathcal{D}$
\WHILE{training not converged}
    \FOR{each environment instance}
        \STATE Randomize domain (dynamics, terrain, kinematics, etc.)
        \STATE Randomize agent commands and robot state $x_0$
        \STATE Reset episode buffer
        \FOR{$t = 0$ to $T_{\text{episode}}$}
        \IF{$t > H$}
            \STATE Estimate current state history $\bar X_{t|t}$ via $\bar{\mu}_\zeta$
            \STATE Solve MPC (Eq.~\eqref{eq: rollout OCP}) via our generalized Gauss--Newton solver (Section~\ref{sec: dynamics and estimation}) and $\hat{\mu}_\theta$
            \STATE Set $u_t$ to first element in optimal action trajectory
        \ELSE
            \STATE Set $u_t$ to nominal action (Appendix~\ref{ap: quadruped experiments})
        \ENDIF
        \STATE Step simulator via $u_t$, and $A_\psi$ (Appendix~\ref{ap: quadruped experiments}) for $x_{t+1}$, $y_{t+1}$ 
        \STATE Store $(x_t, u_t, y_t)$ into replay buffer $\mathcal{D}$
        \ENDFOR
    \ENDFOR
    \FOR{each training iteration}
        \STATE Sample trajectory batch $\{x_t, u_t, y_t\}_{t=0}^{T+H} \sim \mathcal{D}$
        \STATE Update $\theta$ via Algorithm~\ref{alg:dynamics step}
        \STATE Update $\zeta$ via Algorithm~\ref{alg:estimator step}
    \ENDFOR
\ENDWHILE
\STATE Deploy MPC with learned dynamics and estimation in new environments with new costs and constraints
\end{algorithmic}
\end{algorithm}

\begin{algorithm}[t]
\caption{Dynamics training step}
\label{alg:dynamics step}
\footnotesize
\begin{algorithmic}[1]
\REQUIRE Trajectory batch $\{x_t, u_t, y_t\}_{t=0}^{T+H} \sim \mathcal{D}$, initial state estimate covariance $\Sigma_{\bar X}$, measurement noise, dynamics smoothness budget $\hat c_{\text{ub}}$, horizon $T$, history length $H$, SNS dynamics $\hat{\mu}_\theta$, SNS estimator $\bar \mu_\zeta$, discount factor $\gamma$, learning rate, loss weights $\lambda_0$, $\lambda_1$, $\lambda_2$, and $\lambda_3$
    \STATE Perturb initial state history estimate $\bar X_{t|t} \sim \mathcal{N}(X_t, \Sigma_{\bar X})$
    \FOR{$t = H$ to $T+H$}
        \STATE Predict $x_{t+1}$ via $\hat{\mu}_\theta$ with $X_{t}$, $U_{t}$ for $\mathcal{L}_{\text{step}}$ (Eq.~\eqref{eq: L_step})
        \IF{$t > H$}
        \STATE Predict $x_{t+1}$ via $\hat{\mu}_\theta$ with $\hat X_{t}$, $U_{t}$ for $\mathcal{L}_{\text{rollout}}$ (Eq.~\eqref{eq: L_rollout})
        \STATE Predict $x_{t+1}$ via $\hat{\mu}_\theta$ with $\texttt{sg}(\bar X_{t|t})$, $U_{t}$
        for $\mathcal{L}_{\text{corrupt}}$ (Eq.~\eqref{eq: L_corrupt}) and prior $\bar{X}_{t+1|t}$ (Eq.~\eqref{eq:X_bar_predict})
        \ENDIF
        \STATE Corrupt $y_t$ with noise
        \STATE Compute $\nu_{t+1}$ (Eq.~\eqref{eq:innov}) and $g_{\nu,t+1}$ (Eq.~\eqref{eq:innov_grad_exact})
        \STATE Estimate $\bar{X}_{t+1|t+1}$ via $\bar \mu_\zeta$ with $\bar{X}_{t+1|t}$, $U_t$, $Y_{t+1}$, $\nu_{t+1}$, $g_{\nu,t+1}$  
    \ENDFOR
    \STATE Update $\theta$ using total SNS dynamics loss (Eq.~\eqref{eq: dynamics loss})
\end{algorithmic}
\end{algorithm}

\begin{algorithm}[t]
\caption{Estimator training step}
\label{alg:estimator step}
\footnotesize
\begin{algorithmic}[1]
\REQUIRE Trajectory batch $\{x_t, u_t, y_t\}_{t=0}^{T+H} \sim \mathcal{D}$, Initial state estimate covariance $\Sigma_{\bar X}$, measurement noise, estimator smoothness budget $\bar c_{\text{ub}}$, horizon $T$, history length $H$, SNS dynamics $\hat{\mu}_\theta$, SNS estimator $\bar \mu_\zeta$, learning rate, loss weight $\lambda_4$
    \STATE Perturb initial state history estimate $\bar X_{t|t} \sim \mathcal{N}(X_t, \Sigma_{\bar X})$
    \FOR{$t = H+1$ to $T+H$}
        \STATE Predict $x_{t+1}$ via $\hat{\mu}_\theta$ with $\texttt{sg}(\bar X_{t|t})$, $U_{t}$
        for prior $\bar{X}_{t+1|t}$ (Eq.~\eqref{eq:X_bar_predict})
        \STATE Corrupt $y_t$ with noise
        \STATE Compute $\nu_{t+1}$ (Eq.~\eqref{eq:innov}) and $g_{\nu,t}$ (Eq.~\eqref{eq:innov_grad_exact})
        \STATE Estimate $\bar{X}_{t+1|t+1}$ via $\bar \mu_\zeta$ with $\bar{X}_{t+1|t}$, $U_t$, $Y_{t+1}$, $\nu_{t+1}$, $g_{\nu,t+1}$  
    \ENDFOR
    \STATE Update $\zeta$ using total SNS estimator loss (Eq.~\eqref{eq: estimator loss})
\end{algorithmic}
\end{algorithm}

\noindent Unless otherwise noted, all settings described below are shared across the quadruped experiments.

\noindent\textbf{Simulation environments.}
We run MuJoCo simulation at 200\,Hz and perform control and data collection at
50\,Hz. We keep MuJoCo’s default contact settings, including
\texttt{impratio}$=1$, which is known to produce stiff contacts and large
derivatives in legged locomotion~\cite{zhang2025whole}. These settings ensure
that our training data contain the nonsmooth behaviors characteristic of contact
dynamics.

At every reset, we randomize physical and terrain parameters as summarized in
Table~\ref{tab:dr-robot}, including link masses, centers of mass, joint
properties, friction coefficients, foot geometry, and measurement noise. The
ground-plane slope is randomized by sampling a uniform tilt axis and angle. To
deploy for actuator system identification, we train initial models using
MuJoCo’s position-actuator interface with gains
$K_p \sim \mathcal{U}(23,\,27)$\,N·m/rad and
$K_d \sim \mathcal{U}(2.5,\,3.5)$\,N·m·s/rad. All collision geometries in the XML
are converted to capsules or spheres to maintain accurate signed-distance
computations in our MuJoCo version.

\begin{table}[t]
\caption{Domain Randomization Distributions for Go2 Quadruped}
\label{tab:dr-robot}
\vspace{-5pt}
\centering
\begin{tabular}{ll}
\hline
\textbf{Quantity} & \textbf{Distribution} \\
\hline
Link masses $m$ [kg] & $\mathcal{U}(0.975\,m_0,\,1.025\,m_0)$ \\
Base COM offset [mm] & $\mathcal{U}(\pm3)$ per axis \\
Link COM offsets [mm] & $\mathcal{U}(\pm1)$ per axis \\
Joint damping [N·m·s/rad] & $\mathcal{U}(0.0,\,0.05)$ \\
Joint friction [N·m·s/rad] & $\mathcal{U}(0.0,\,0.25)$ \\
Joint armature [kg·m$^2$] & $\mathcal{U}(0,\,5{\times}10^{-5})$ \\
Foot radius $r$ [m] & $\mathcal{U}(0.95\,r_0,\,1.05\,r_0)$ \\
Sliding friction $\mu_{\mathrm{slide}}$ & $\mathcal{U}(0.2,\,1.0)$ \\
Rolling/torsional friction & Scaled $\mu_{\mathrm{slide}}$, clipped to $[0.1,\,1.0]$ \\
Terrain tilt axis & Uniform on unit sphere \\
Terrain tilt angle [rad] & $\mathcal{U}(0,\,0.5)$ \\
Latency [ms] & $\mathcal{U}\{10,15\}$ \\
Measurement noise: & \\
\quad $r^{6D}_\mathcal{W}$ & $\mathcal{U}(\pm0.001)$ \\
\quad $q_J$ [rad] & $\mathcal{U}(\pm0.01)$ \\
\quad $v_J$ [rad/s] & $\mathcal{U}(\pm0.1)$ \\
\quad $\dot v_\mathcal{B}$ [m/s$^2$] & $\mathcal{U}(\pm0.08)$ \\
\quad $\omega_\mathcal{B}$ [rad/s] & $\mathcal{U}(\pm0.025)$ \\
\hline
\end{tabular}
\end{table}

\noindent\textbf{Data collection.}
Each episode lasts 5.12\,s (256 control steps), and we simulate 512 environments
in parallel. The replay buffer stores 20,480 episodes
($\approx$5M transitions) and cycles in a FIFO manner. Reference commands are
varied across episodes, and robot states are randomized at every reset. We collect
an initial bootstrap buffer using random uniform spline actions between the
minimum and maximum action limits, then alternate between collecting on-policy
trajectories and updating our models using the replay buffer. The test dataset
(Table~\ref{tab:MAE}) is generated using the same procedure.

\noindent\textbf{Hybrid simulation and actuator models.}
Our control pipeline uses hybrid simulation that combines MuJoCo rigid-body
dynamics with a smooth neural surrogate actuator model. For all 12 joints, we
log torque, commanded position, actual position, and velocity at 200\,Hz.
Following~\cite{margolis2023walk, lee2020learning}, we train
actuator networks and introduce randomized latency. An actuator network is also used to accurately capture mechanical work and torque penalties (Table~\ref{tab:costs}) during MPC rollouts.


\noindent\textbf{Model training setup.}
Unless otherwise noted, all learned models use Mish activations and are updated 500 times between data-collection episodes using the Lion optimizer~\cite{chen2023symbolic}. The
global loss uses weights $\lambda_0=\lambda_1=0.5$, $\lambda_2=0.05$, and
smoothness penalties $\lambda_3=10.0$, $\lambda_4=10^{-5}$. SNS models
use sensitivity budgets of $10^4$ for dynamics and~1 for the estimator. The
estimator noise covariance is initialized as
$\tfrac12 \Sigma_{\mathcal{D}}$, computed from the initial replay buffer. All
models use history length $H=8$, rollout horizon $T=19$, and batch size 512
trajectories (Algorithms~\ref{alg:dynamics step}--\ref{alg:estimator step}). All simulation experiments are conducted on a Linux server with 256 CPUs and 8 GPUs. Individual experiments use CPU-based multi-threading for MuJoCo and a single GPU per experiment. Checkpoints are saved every 500 training steps.

\noindent\textbf{Network architectures and learning rates.}
For simulation experiments, the dynamics / estimator learning rates are 0.0008 / 0.0004 for Cauchy-loss SNS models, 0.0004 / 0.0002 for Gaussian-loss SNS models, and 0.0001 / 0.00005 for standard and weight-decay MLPs.
These models use four hidden layers with widths $2x_0$ (dynamics) and
$1.5y_0$ (estimator), where $x_0$ and $y_0$ denote input dimensions. All models use history length $H=8$. For hardware deployment, we use the same learning rates with four-layer MLPs
with widths $[1.5x_0,\,1.5x_0,\,1.25x_0,\,1.0x_0]$ (dynamics) and $[1.5y_0,\,1.25y_0,\,1.0y_0,\,0.75y_0]$ (estimator).
These architectures incorporate the same smooth-surrogate constraints, and the
same trained model is used across all real-world experiments.

\noindent\textbf{Costs and constraints.}
We use quadratic tracking objectives during training, validation, and
deployment, with task-specific weights listed in Table~\ref{tab:costs}. The cost
penalizes deviations in base and joint motion, torque effort, gait phase and foot
height, and includes a drift term to reduce long-horizon bias.

Bounding and galloping enforce
collision-avoidance constraints on the thigh and upper shank,
$\phi_{\text{thigh}}, \phi_{\text{shank}} > 0$, with relaxation
$\delta = 5\cdot 10^{-4}$ and coefficient $\beta = 10^{-3}$. Tripod additionally
enforces base-height bounds $-0.02 < \varepsilon_z < 0.08$\,m, with
$\delta=0.01$ and $\beta = 1.0$.

\noindent\textbf{Behavior stitching and transitions.}
Costs, constraints, and commands are regenerated at each time step and appended
to the end of the planning horizon, making the trajectory optimization
time-varying. Directly applying new gait parameters can lead to jerky motions or
catastrophic failures due to poor foot-height commands. To mitigate this, a
transition-specific cost (Table~\ref{tab:costs}) is activated whenever the cost
definition changes, temporarily suppressing gait-specific tracking and allowing
the optimizer to discover a valid transition. All commands and weights are
smoothed using an exponential moving average before being appended to the
horizon. This mechanism is only used during evaluation and deployment.

\noindent\textbf{Reference commands.}
A periodic gait generator produces foot-height references
$\phi^{\mathrm{cmd}}_{\mathrm{foot}}$ parameterized by cadence, phase, duty ratio, and swing height~\cite{howell2022predictive}. In tripod mode, the disabled leg is assigned a fixed 0.1\,m foot-height command; during rearing, both rear legs are disabled. Base height varies throughout training and deployment with nominal height $z^{\mathrm{cmd}} = 0.27$. Nominal joint positions are
$q_J^{\mathrm{nom}} = [0,\,0.9,\,-1.8]\times 4$. Training references are randomized using three distributions. Each batch of episodes contains an equal mixture of: (i) \emph{full random commands:}
$(v_x, v_y) \sim \mathcal{U}(-2.0,\,2.0)$\,m/s,
$\omega_{\mathrm{yaw}} \sim \mathcal{U}(-\pi,\,\pi)$\,rad/s,
$z \sim \mathcal{U}(0.06,\,0.8)$\,m,
roll, pitch, yaw $\sim \mathcal{U}(-\pi,\,\pi)$; (ii) \emph{random locomotion commands:}
$(v_x, v_y, \omega_{\mathrm{yaw}})$ as above,
$z \sim \mathcal{U}(0.2,\,0.35)$\,m,
roll and pitch fixed to zero; (iii) \emph{random pose commands:} zero velocities,
$z \sim \mathcal{U}(0.35,\,0.55)$\,m,
roll $\sim \mathcal{U}(-\pi/4,\,\pi/4)$,
pitch $\sim \mathcal{U}(-\pi/2,\,\pi/2)$,
yaw $\sim \mathcal{U}(-\pi,\,\pi)$.

\noindent\textbf{Section-specific experiment settings.} Results are averaged across either 10 random seeds (Sec.~\ref{sec: convergence} and~\ref{sec: unlocking gradient-based}), or 5 random seeds (Sec.~\ref{sec: zero-shot}) governing the initial robot state and, for sampling-based methods, control samples.

\emph{Convergence experiments (Sec.~\ref{sec: convergence}).}
Linear splines with $k=5$ knots and 16 rollouts are used with GGN-MPC.
Fig.~\ref{fig: mse vs cauchy} displays results across four
independently trained networks per configuration.

\emph{Zero-shot generalization (Sec.~\ref{sec: zero-shot}).}
Four models (different seeds) are trained to completion for each network
configuration. Models shown in Tables~\ref{tab:MAE} and
\ref{tab:performance_methods} are selected based on trotting performance at the
end of training. The early-stopped model is frozen at the 12,500-step checkpoint.
For Fig.~\ref{fig: traj compare}, and
Table~\ref{tab:performance_methods}, we use $k=6$ knots and 16 rollouts.

\emph{Gradient-based MPC comparisons (Sec.~\ref{sec: unlocking gradient-based}).}
The same SNS-Cauchy models from Sec.~\ref{sec: zero-shot} are used to evaluate
GGN-MPC and DIAL-MPC across five trotting tasks (identical to
Sec.~\ref{sec: convergence}).

\emph{Parameterization comparison (Sec.~\ref{sec: convergence}, Fig.~\ref{fig: lip vs sns mpc}).}
As in Appendix~\ref{ap: sns experiments}, SNS and Lipschitz MLP dynamics are trained with Adam ($\beta_1=0.9$, $\beta_2=0.999$), with shared dynamics learning rates of $8\cdot10^{-4}$ (smaller width) and $4\cdot10^{-4}$ (larger width). Both widths use five hidden layers, with widths $[2,2,2,2,1]\,x_0$ ($\approx$5.6M parameters) and $4x_0$ ($\approx$23.7M parameters), where $x_0=588$. Every checkpoint (every 500 steps) is evaluated on the five trotting tasks with 10 seeds per task. All other settings are as above.

\noindent\textbf{Hardware deployment.}
All deployment algorithms run on a Linux workstation with an NVIDIA RTX~4090 GPU and
are deployed to the Unitree Go2 via a wired connection. To convert MPC outputs
into joint torques, we combine impedance and feedforward terms:
\begin{equation}
\begin{aligned}
\tau
&= K_p (q^d_J - q_J) - K_d v_J \\
&\quad + K^{\mathrm{ff}}_p (q^*_{J,t+1} - q_J)
      - K^{\mathrm{ff}}_d (v^*_{J,t+1} - v_J),
\end{aligned}
\end{equation}
where starred quantities denote MPC-predicted next-step joint positions and
velocities. Deployment uses linear splines with $k=6$ knots and GGN-MPC with
8 rollouts.

\bibliography{references}

\begin{thebibliography}{10}
\providecommand{\url}[1]{#1}
\csname url@samestyle\endcsname
\providecommand{\newblock}{\relax}
\providecommand{\bibinfo}[2]{#2}
\providecommand{\BIBentrySTDinterwordspacing}{\spaceskip=0pt\relax}
\providecommand{\BIBentryALTinterwordstretchfactor}{4}
\providecommand{\BIBentryALTinterwordspacing}{\spaceskip=\fontdimen2\font plus
\BIBentryALTinterwordstretchfactor\fontdimen3\font minus
  \fontdimen4\font\relax}
\providecommand{\BIBforeignlanguage}[2]{{%
\expandafter\ifx\csname l@#1\endcsname\relax
\typeout{** WARNING: IEEEtran.bst: No hyphenation pattern has been}%
\typeout{** loaded for the language `#1'. Using the pattern for}%
\typeout{** the default language instead.}%
\else
\language=\csname l@#1\endcsname
\fi
#2}}
\providecommand{\BIBdecl}{\relax}
\BIBdecl

\bibitem{lee2020learning}
J.~Lee, J.~Hwangbo, L.~Wellhausen, V.~Koltun, and M.~Hutter, ``Learning
  quadrupedal locomotion over challenging terrain,'' \emph{Science robotics},
  vol.~5, no.~47, p. eabc5986, 2020.

\bibitem{kumar2021rma}
A.~Kumar, Z.~Fu, D.~Pathak, and J.~Malik, ``Rma: Rapid motor adaptation for
  legged robots,'' \emph{arXiv preprint arXiv:2107.04034}, 2021.

\bibitem{tassa2012synthesis}
Y.~Tassa, T.~Erez, and E.~Todorov, ``Synthesis and stabilization of complex
  behaviors through online trajectory optimization,'' in \emph{2012 IEEE/RSJ
  International Conference on Intelligent Robots and Systems}.\hskip 1em plus
  0.5em minus 0.4em\relax IEEE, 2012, pp. 4906--4913.

\bibitem{howell2022predictive}
T.~Howell, N.~Gileadi, S.~Tunyasuvunakool, K.~Zakka, T.~Erez, and Y.~Tassa,
  ``Predictive sampling: Real-time behaviour synthesis with mujoco,''
  \emph{arXiv preprint arXiv:2212.00541}, 2022.

\bibitem{lutter2021learning}
M.~Lutter, L.~Hasenclever, A.~Byravan, G.~Dulac-Arnold, P.~Trochim, N.~Heess,
  J.~Merel, and Y.~Tassa, ``Learning dynamics models for model predictive
  agents,'' \emph{arXiv preprint arXiv:2109.14311}, 2021.

\bibitem{posa2013direct}
M.~Posa and R.~Tedrake, ``Direct trajectory optimization of rigid body
  dynamical systems through contact,'' in \emph{Algorithmic Foundations of
  Robotics X: Proceedings of the Tenth Workshop on the Algorithmic Foundations
  of Robotics}.\hskip 1em plus 0.5em minus 0.4em\relax Springer, 2013, pp.
  527--542.

\bibitem{kelly2017introduction}
M.~Kelly, ``An introduction to trajectory optimization: How to do your own
  direct collocation,'' \emph{SIAM review}, vol.~59, no.~4, pp. 849--904, 2017.

\bibitem{kim2025contact}
G.~Kim, D.~Kang, J.-H. Kim, S.~Hong, and H.-W. Park, ``Contact-implicit model
  predictive control: Controlling diverse quadruped motions without pre-planned
  contact modes or trajectories,'' \emph{The International Journal of Robotics
  Research}, vol.~44, no.~3, pp. 486--510, 2025.

\bibitem{liu2022learning}
H.-T.~D. Liu, F.~Williams, A.~Jacobson, S.~Fidler, and O.~Litany, ``Learning
  smooth neural functions via lipschitz regularization,'' in \emph{ACM SIGGRAPH
  2022 Conference Proceedings}, 2022, pp. 1--13.

\bibitem{rosca2020case}
\BIBentryALTinterwordspacing
M.~Rosca, T.~Weber, A.~Gretton, and S.~Mohamed, ``A case for new neural network
  smoothness constraints,'' in \emph{Proceedings on "I Can't Believe It's Not
  Better!" at NeurIPS Workshops}, ser. Proceedings of Machine Learning
  Research, J.~Zosa~Forde, F.~Ruiz, M.~F. Pradier, and A.~Schein, Eds., vol.
  137.\hskip 1em plus 0.5em minus 0.4em\relax PMLR, 12 Dec 2020, pp. 21--32.
  [Online]. Available: \url{https://proceedings.mlr.press/v137/rosca20a.html}
\BIBentrySTDinterwordspacing

\bibitem{anil2019sorting}
C.~Anil, J.~Lucas, and R.~Grosse, ``Sorting out lipschitz function
  approximation,'' in \emph{International conference on machine
  learning}.\hskip 1em plus 0.5em minus 0.4em\relax PMLR, 2019, pp. 291--301.

\bibitem{gulrajani2017improved}
I.~Gulrajani, F.~Ahmed, M.~Arjovsky, V.~Dumoulin, and A.~C. Courville,
  ``Improved training of wasserstein gans,'' \emph{Advances in neural
  information processing systems}, vol.~30, 2017.

\bibitem{yoshida2017spectral}
Y.~Yoshida and T.~Miyato, ``Spectral norm regularization for improving the
  generalizability of deep learning,'' \emph{arXiv preprint arXiv:1705.10941},
  2017.

\bibitem{song2023lipsnet}
X.~Song, J.~Duan, W.~Wang, S.~E. Li, C.~Chen, B.~Cheng, B.~Zhang, J.~Wei, and
  X.~S. Wang, ``Lipsnet: a smooth and robust neural network with adaptive
  lipschitz constant for high accuracy optimal control,'' in
  \emph{International Conference on Machine Learning}.\hskip 1em plus 0.5em
  minus 0.4em\relax PMLR, 2023, pp. 32\,253--32\,272.

\bibitem{tan2024robust}
W.~G.~Y. Tan and Z.~Wu, ``Robust machine learning modeling for predictive
  control using lipschitz-constrained neural networks,'' \emph{Computers \&
  Chemical Engineering}, vol. 180, p. 108466, 2024.

\bibitem{xu2025neural}
J.~Xu, E.~Heiden, I.~Akinola, D.~Fox, M.~Macklin, and Y.~Narang, ``Neural robot
  dynamics,'' \emph{arXiv preprint arXiv:2508.15755}, 2025.

\bibitem{chua2018deep}
K.~Chua, R.~Calandra, R.~McAllister, and S.~Levine, ``Deep reinforcement
  learning in a handful of trials using probabilistic dynamics models,''
  \emph{Advances in neural information processing systems}, vol.~31, 2018.

\bibitem{amigo2025first}
J.~Amigo, R.~Khorrambakht, E.~Chane-Sane, N.~Mansard, and L.~Righetti, ``First
  order model-based rl through decoupled backpropagation,'' \emph{arXiv
  preprint arXiv:2509.00215}, 2025.

\bibitem{byravan2021evaluating}
A.~Byravan, L.~Hasenclever, P.~Trochim, M.~Mirza, A.~D. Ialongo, Y.~Tassa,
  J.~T. Springenberg, A.~Abdolmaleki, N.~Heess, J.~Merel, and M.~Riedmiller,
  ``Evaluating model-based planning and planner amortization for continuous
  control,'' \emph{arXiv preprint arXiv:2110.03363}, 2021.

\bibitem{parmar2021fundamental}
M.~Parmar, M.~Halm, and M.~Posa, ``Fundamental challenges in deep learning for
  stiff contact dynamics,'' in \emph{2021 IEEE/RSJ International Conference on
  Intelligent Robots and Systems (IROS)}.\hskip 1em plus 0.5em minus
  0.4em\relax IEEE, 2021, pp. 5181--5188.

\bibitem{roth2025learned}
P.~Roth, J.~Frey, C.~Cadena, and M.~Hutter, ``Learned perceptive forward
  dynamics model for safe and platform-aware robotic navigation,'' \emph{arXiv
  preprint arXiv:2504.19322}, 2025.

\bibitem{tsakalides1994maximum}
P.~Tsakalides and C.~Nikias, ``Maximum likelihood localization of sources in
  noise modeled as a cauchy process,'' in \emph{Proceedings of
  MILCOM'94}.\hskip 1em plus 0.5em minus 0.4em\relax IEEE, 1994, pp. 613--617.

\bibitem{nocedal2006numerical}
J.~Nocedal and S.~J. Wright, \emph{Numerical optimization}.\hskip 1em plus
  0.5em minus 0.4em\relax Springer, 2006.

\bibitem{kelly2017transcription}
M.~P. Kelly, ``Transcription methods for trajectory optimization: a beginners
  tutorial,'' \emph{arXiv preprint arXiv:1707.00284}, 2017.

\bibitem{xue2025full}
H.~Xue, C.~Pan, Z.~Yi, G.~Qu, and G.~Shi, ``Full-order sampling-based mpc for
  torque-level locomotion control via diffusion-style annealing,'' in
  \emph{2025 IEEE International Conference on Robotics and Automation
  (ICRA)}.\hskip 1em plus 0.5em minus 0.4em\relax IEEE, 2025, pp. 4974--4981.

\bibitem{lee2025sym2real}
E.~Lee, S.~A. Moore, and B.~Chen, ``Sym2real: Symbolic dynamics with residual
  learning for data-efficient adaptive control,'' \emph{arXiv preprint
  arXiv:2509.15412}, 2025.

\bibitem{nagabandi2020deep}
A.~Nagabandi, K.~Konolige, S.~Levine, and V.~Kumar, ``Deep dynamics models for
  learning dexterous manipulation,'' in \emph{Conference on robot
  learning}.\hskip 1em plus 0.5em minus 0.4em\relax PMLR, 2020, pp. 1101--1112.

\bibitem{alvarez2025real}
J.~Alvarez-Padilla, J.~Z. Zhang, S.~Kwok, J.~M. Dolan, and Z.~Manchester,
  ``Real-time whole-body control of legged robots with model-predictive path
  integral control,'' in \emph{2025 IEEE International Conference on Robotics
  and Automation (ICRA)}.\hskip 1em plus 0.5em minus 0.4em\relax IEEE, 2025,
  pp. 14\,721--14\,727.

\bibitem{sun2018evolving}
Y.~Sun, G.~G. Yen, and Z.~Yi, ``Evolving unsupervised deep neural networks for
  learning meaningful representations,'' \emph{IEEE Transactions on
  Evolutionary Computation}, vol.~23, no.~1, pp. 89--103, 2018.

\bibitem{wensing2024optimization}
P.~M. Wensing, M.~Posa, Y.~Hu, A.~Escande, N.~Mansard, and A.~Del~Prete,
  ``Optimization-based control for dynamic legged robots,'' \emph{IEEE
  Transactions on Robotics}, vol.~40, pp. 43--63, 2024.

\bibitem{le2024fast}
S.~Le~Cleac'h, T.~A. Howell, S.~Yang, C.-Y. Lee, J.~Zhang, A.~Bishop,
  M.~Schwager, and Z.~Manchester, ``Fast contact-implicit model predictive
  control,'' \emph{IEEE Transactions on Robotics}, vol.~40, pp. 1617--1629,
  2024.

\bibitem{zhang2025whole}
J.~Z. Zhang, T.~A. Howell, Z.~Yi, C.~Pan, G.~Shi, G.~Qu, T.~Erez, Y.~Tassa, and
  Z.~Manchester, ``Whole-body model-predictive control of legged robots with
  mujoco,'' \emph{arXiv preprint arXiv:2503.04613}, 2025.

\bibitem{mordatch2012discovery}
I.~Mordatch, E.~Todorov, and Z.~Popovi{\'c}, ``Discovery of complex behaviors
  through contact-invariant optimization,'' \emph{ACM Transactions on Graphics
  (ToG)}, vol.~31, no.~4, pp. 1--8, 2012.

\bibitem{erez2012trajectory}
T.~Erez and E.~Todorov, ``Trajectory optimization for domains with contacts
  using inverse dynamics,'' in \emph{2012 IEEE/RSJ International conference on
  intelligent robots and systems}.\hskip 1em plus 0.5em minus 0.4em\relax IEEE,
  2012, pp. 4914--4919.

\bibitem{khazoom2024tailoring}
C.~Khazoom, S.~Hong, M.~Chignoli, E.~Stanger-Jones, and S.~Kim, ``Tailoring
  solution accuracy for fast whole-body model predictive control of legged
  robots,'' \emph{IEEE Robotics and Automation Letters}, vol.~9, no.~12, pp.
  11\,074--11\,081, 2024.

\bibitem{li2025cafempc}
H.~Li and P.~M. Wensing, ``{CAFE-MPC}: A cascaded-fidelity model predictive
  control framework with tuning-free whole-body control,'' \emph{IEEE
  Transactions on Robotics}, vol.~41, 2025.

\bibitem{ponton2021efficient}
B.~Ponton, M.~Khadiv, A.~Meduri, and L.~Righetti, ``Efficient multicontact
  pattern generation with sequential convex approximations of the centroidal
  dynamics,'' \emph{IEEE Transactions on Robotics}, vol.~37, no.~5, pp.
  1661--1679, 2021.

\bibitem{meduri2023biconmp}
A.~Meduri, P.~Shah, J.~Viereck, M.~Khadiv, I.~Havoutis, and L.~Righetti,
  ``{BiConMP}: A nonlinear model predictive control framework for whole body
  motion planning,'' \emph{IEEE Transactions on Robotics}, vol.~39, no.~2, pp.
  905--922, 2023.

\bibitem{howell2022dojo}
T.~A. Howell, S.~Le~Cleac’h, J.~Z. Kolter, M.~Schwager, and Z.~Manchester,
  ``Dojo: A differentiable simulator for robotics,'' \emph{arXiv preprint
  arXiv:2203.00806}, vol.~9, no.~2, p.~4, 2022.

\bibitem{suh2022bundled}
H.~J.~T. Suh, T.~Pang, and R.~Tedrake, ``Bundled gradients through contact via
  randomized smoothing,'' \emph{IEEE Robotics and Automation Letters}, vol.~7,
  no.~2, pp. 4000--4007, 2022.

\bibitem{pang2023global}
T.~Pang, H.~T. Suh, L.~Yang, and R.~Tedrake, ``Global planning for contact-rich
  manipulation via local smoothing of quasi-dynamic contact models,''
  \emph{IEEE Transactions on robotics}, vol.~39, no.~6, pp. 4691--4711, 2023.

\bibitem{xu2022accelerated}
J.~Xu, V.~Makoviychuk, Y.~Narang, F.~Ramos, W.~Matusik, A.~Garg, and
  M.~Macklin, ``Accelerated policy learning with parallel differentiable
  simulation,'' \emph{arXiv preprint arXiv:2204.07137}, 2022.

\bibitem{georgiev2024adaptive}
I.~Georgiev, K.~Srinivasan, J.~Xu, E.~Heiden, and A.~Garg, ``Adaptive horizon
  actor-critic for policy learning in contact-rich differentiable simulation,''
  \emph{arXiv preprint arXiv:2405.17784}, 2024.

\bibitem{schwarke2025learning}
C.~Schwarke, V.~Klemm, J.~Bagajo, J.~P. Sleiman, I.~Georgiev, J.~T. Torres, and
  M.~Hutter, ``Learning deployable locomotion control via differentiable
  simulation,'' in \emph{9th Annual Conference on Robot Learning}, 2025.

\bibitem{hafner2019dream}
D.~Hafner, T.~Lillicrap, J.~Ba, and M.~Norouzi, ``Dream to control: Learning
  behaviors by latent imagination,'' \emph{arXiv preprint arXiv:1912.01603},
  2019.

\bibitem{li2025robotic}
C.~Li, A.~Krause, and M.~Hutter, ``Robotic world model: A neural network
  simulator for robust policy optimization in robotics,'' \emph{arXiv preprint
  arXiv:2501.10100}, 2025.

\bibitem{williams2017information}
G.~Williams, N.~Wagener, B.~Goldfain, P.~Drews, J.~M. Rehg, B.~Boots, and E.~A.
  Theodorou, ``Information theoretic mpc for model-based reinforcement
  learning,'' in \emph{2017 IEEE international conference on robotics and
  automation (ICRA)}.\hskip 1em plus 0.5em minus 0.4em\relax IEEE, 2017, pp.
  1714--1721.

\bibitem{hu2025egocentric}
Y.~Hu, B.~Chen, and H.~Lipson, ``Egocentric visual self-modeling for autonomous
  robot dynamics prediction and adaptation,'' \emph{npj Robotics}, vol.~3,
  no.~1, p.~14, 2025.

\bibitem{janner2022planning}
M.~Janner, Y.~Du, J.~B. Tenenbaum, and S.~Levine, ``Planning with diffusion for
  flexible behavior synthesis,'' \emph{arXiv preprint arXiv:2205.09991}, 2022.

\bibitem{huang2025flexible}
R.~Huang, H.~Balim, H.~Yang, and Y.~Du, ``Flexible locomotion learning with
  diffusion model predictive control,'' \emph{arXiv preprint arXiv:2510.04234},
  2025.

\bibitem{li2025bfmzero}
Y.~Li, Z.~Luo, T.~Zhang, C.~Dai, A.~Kanervisto, A.~Tirinzoni, H.~Weng,
  K.~Kitani, M.~Guzek, A.~Touati, A.~Lazaric, M.~Pirotta, and G.~Shi,
  ``{BFM-Zero}: A promptable behavioral foundation model for humanoid control
  using unsupervised reinforcement learning,'' 2025.

\bibitem{pfrommer2021contactnets}
S.~Pfrommer, M.~Halm, and M.~Posa, ``Contactnets: Learning discontinuous
  contact dynamics with smooth, implicit representations,'' in \emph{Conference
  on Robot Learning}.\hskip 1em plus 0.5em minus 0.4em\relax PMLR, 2021, pp.
  2279--2291.

\bibitem{hochlehnert2021learning}
A.~Hochlehnert, A.~Terenin, S.~S{\ae}mundsson, and M.~Deisenroth, ``Learning
  contact dynamics using physically structured neural networks,'' in
  \emph{International Conference on Artificial Intelligence and
  Statistics}.\hskip 1em plus 0.5em minus 0.4em\relax PMLR, 2021, pp.
  2152--2160.

\bibitem{lutter2019deep}
M.~Lutter, C.~Ritter, and J.~Peters, ``Deep lagrangian networks: Using physics
  as model prior for deep learning,'' \emph{arXiv preprint arXiv:1907.04490},
  2019.

\bibitem{sukhija2022gradient}
B.~Sukhija, N.~K{\"o}hler, M.~Zamora, S.~Zimmermann, S.~Curi, A.~Krause, and
  S.~Coros, ``Gradient-based trajectory optimization with learned dynamics,''
  \emph{arXiv preprint arXiv:2204.04558}, 2022.

\bibitem{heiden2021neuralsim}
E.~Heiden, D.~Millard, E.~Coumans, Y.~Sheng, and G.~S. Sukhatme, ``Neuralsim:
  Augmenting differentiable simulators with neural networks,'' in \emph{2021
  IEEE International Conference on Robotics and Automation (ICRA)}.\hskip 1em
  plus 0.5em minus 0.4em\relax IEEE, 2021, pp. 9474--9481.

\bibitem{miyato2018spectral}
T.~Miyato, T.~Kataoka, M.~Koyama, and Y.~Yoshida, ``Spectral normalization for
  generative adversarial networks,'' \emph{arXiv preprint arXiv:1802.05957},
  2018.

\bibitem{gouk2021regularisation}
H.~Gouk, E.~Frank, B.~Pfahringer, and M.~J. Cree, ``Regularisation of neural
  networks by enforcing lipschitz continuity,'' \emph{Machine Learning}, vol.
  110, no.~2, pp. 393--416, 2021.

\bibitem{qi2023lipsformer}
X.~Qi, J.~Wang, Y.~Chen, Y.~Shi, and L.~Zhang, ``Lipsformer: Introducing
  lipschitz continuity to vision transformers,'' \emph{arXiv preprint
  arXiv:2304.09856}, 2023.

\bibitem{mlotshwa2022cauchy}
T.~Mlotshwa, H.~van Deventer, and A.~S. Bosman, ``Cauchy loss function:
  Robustness under gaussian and cauchy noise,'' in \emph{Southern African
  Conference for Artificial Intelligence Research}.\hskip 1em plus 0.5em minus
  0.4em\relax Springer, 2022, pp. 123--138.

\bibitem{el2009robust}
M.~T. El-Melegy, M.~H. Essai, and A.~A. Ali, ``Robust training of artificial
  feedforward neural networks,'' in \emph{Foundations of Computational,
  Intelligence Volume 1: Learning and Approximation}.\hskip 1em plus 0.5em
  minus 0.4em\relax Springer, 2009, pp. 217--242.

\bibitem{barron2019general}
J.~T. Barron, ``A general and adaptive robust loss function,'' in
  \emph{Proceedings of the IEEE/CVF conference on computer vision and pattern
  recognition}, 2019, pp. 4331--4339.

\bibitem{moore2025automated}
S.~A. Moore, B.~P. Mann, and B.~Chen, ``Automated global analysis of
  experimental dynamics through low-dimensional linear embeddings,'' \emph{npj
  Complexity}, vol.~2, no.~1, p.~36, 2025.

\bibitem{kim2021legged}
J.-H. Kim, S.~Hong, G.~Ji, S.~Jeon, J.~Hwangbo, J.-H. Oh, and H.-W. Park,
  ``Legged robot state estimation with dynamic contact event information,''
  \emph{IEEE Robotics and Automation Letters}, vol.~6, no.~4, pp. 6733--6740,
  2021.

\bibitem{hartley2020contact}
R.~Hartley, M.~Ghaffari, R.~M. Eustice, and J.~W. Grizzle, ``Contact-aided
  invariant extended kalman filtering for robot state estimation,'' \emph{The
  International Journal of Robotics Research}, vol.~39, no.~4, pp. 402--430,
  2020.

\bibitem{ji2022concurrent}
G.~Ji, J.~Mun, H.~Kim, and J.~Hwangbo, ``Concurrent training of a control
  policy and a state estimator for dynamic and robust legged locomotion,''
  \emph{IEEE Robotics and Automation Letters}, vol.~7, no.~2, pp. 4630--4637,
  2022.

\bibitem{kim2025learning}
G.~Kim, Y.-H. Lee, and H.-W. Park, ``A learning framework for diverse legged
  robot locomotion using barrier-based style rewards,'' in \emph{2025 IEEE
  International Conference on Robotics and Automation (ICRA)}.\hskip 1em plus
  0.5em minus 0.4em\relax IEEE, 2025, pp. 10\,004--10\,010.

\bibitem{revach2022kalmannet}
G.~Revach, N.~Shlezinger, X.~Ni, A.~L. Escoriza, R.~J. Van~Sloun, and Y.~C.
  Eldar, ``Kalmannet: Neural network aided kalman filtering for partially known
  dynamics,'' \emph{IEEE Transactions on Signal Processing}, vol.~70, pp.
  1532--1547, 2022.

\bibitem{abarbanel2012analysis}
H.~Abarbanel, \emph{Analysis of observed chaotic data}.\hskip 1em plus 0.5em
  minus 0.4em\relax Springer Science \& Business Media, 2012.

\bibitem{schraudolph2002fast}
N.~N. Schraudolph, ``Fast curvature matrix-vector products for second-order
  gradient descent,'' \emph{Neural computation}, vol.~14, no.~7, pp.
  1723--1738, 2002.

\bibitem{grandia2023perceptive}
R.~Grandia, F.~Jenelten, S.~Yang, F.~Farshidian, and M.~Hutter, ``Perceptive
  locomotion through nonlinear model-predictive control,'' \emph{IEEE
  Transactions on Robotics}, vol.~39, no.~5, pp. 3402--3421, 2023.

\bibitem{messerer2021survey}
F.~Messerer, K.~Baumg{\"a}rtner, and M.~Diehl, ``Survey of sequential convex
  programming and generalized gauss-newton methods,'' \emph{ESAIM: Proceedings
  and Surveys}, vol.~71, pp. 64--88, 2021.

\bibitem{todorov2005generalized}
E.~Todorov and W.~Li, ``A generalized iterative lqg method for locally-optimal
  feedback control of constrained nonlinear stochastic systems,'' in
  \emph{Proceedings of the 2005, American Control Conference, 2005.}\hskip 1em
  plus 0.5em minus 0.4em\relax IEEE, 2005, pp. 300--306.

\bibitem{zhou2019continuity}
Y.~Zhou, C.~Barnes, J.~Lu, J.~Yang, and H.~Li, ``On the continuity of rotation
  representations in neural networks,'' in \emph{Proceedings of the IEEE/CVF
  conference on computer vision and pattern recognition}, 2019, pp. 5745--5753.

\bibitem{margolis2023walk}
G.~B. Margolis and P.~Agrawal, ``Walk these ways: Tuning robot control for
  generalization with multiplicity of behavior,'' in \emph{Conference on Robot
  Learning}.\hskip 1em plus 0.5em minus 0.4em\relax PMLR, 2023, pp. 22--31.

\bibitem{chen2023symbolic}
X.~Chen, C.~Liang, D.~Huang, E.~Real, K.~Wang, H.~Pham, X.~Dong, T.~Luong,
  C.-J. Hsieh, Y.~Lu, and Q.~V. Le, ``Symbolic discovery of optimization
  algorithms,'' \emph{Advances in neural information processing systems},
  vol.~36, pp. 49\,205--49\,233, 2023.

\end{thebibliography}
%

\bibliographystyle{IEEEtran}

\end{document}